\RequirePackage{fix-cm}
\documentclass[twocolumn]{svjour3}          % twocolumn
\smartqed  % flush right qed marks, e.g. at end of proof
\usepackage{graphicx}
\usepackage{amsmath}
\usepackage{algorithm}
\usepackage{algorithmic}
\usepackage{makecell}
\usepackage{booktabs}
\usepackage{subfigure}
\usepackage{cite}
\begin{document}

\title{Across-Task Neural Architecture Search via Meta Learning%\thanks{Grants or other notes
}


\author{Jingtao Rong         \and
        Xinyi Yu* \and
        Mingyang Zhang\textsuperscript{$\dagger$} \and
        Linlin Ou* %etc.
}

\institute{Jintao Rong \at
           Department of Information Engineering, Zhejiang University of Technology, Hangzhou 310023, China \\
           \email{2111903071@zjut.edu.cn}           %  \\
           \and
           Linlin Ou \at
           Department of Information Engineering, Zhejiang University of Technology, Hangzhou 310023, China \\
           \email{linlinou@zjut.edu.cn}
           \and
           $\dagger$ Mingyang Zhang contributes equally and share first-authorship
}

\date{Received: date / Accepted: date}

\maketitle

\begin{abstract}
%Insert your abstract here. Include keywords, PACS and mathematical
%subject classification numbers as needed.
  Adequate labeled data and expensive compute resources are the prerequisites for the success of neural architecture search(NAS).
  It is challenging to apply NAS in meta-learning scenarios with limited compute resources and data.
  In this paper, an across-task neural architecture search (AT-NAS) is proposed to address the problem through combining gradient-based meta-learning with EA-based NAS to learn over the distribution of tasks.
  The supernet is learned over an entire set of tasks by meta-learning its weights. Architecture encodes of subnets sampled from the supernet are iteratively adapted by evolutionary algorithms while simultaneously searching for a task-sensitive meta-network. Searched meta-network can be adapted to a novel task via a few learning steps and only costs a little search time.
  Empirical results show that AT-NAS surpasses the related approaches on few-shot classification accuracy. The performance of AT-NAS on classification benchmarks is comparable to that of models searched from scratch, by adapting the architecture in less than an hour from a 5-GPU-day pretrained meta-network.
%  The meta-network can be adapted to a novel task via a few learning steps, only cost a litter search time.
%AT-NAS searches a group of architecture that can be adapted to a novel task via a few learning steps.
%  Therefore, learning task-dependent architecture is more flexible with requiring cheap computation costs and only a little data.
%  Empirical results show that AT-NAS achieves state-of-the-art performance in related few-shot learning approaches.
%  The supervised learning experiments also show that the classification accuracy of AT-NAS is comparable to that of models search from scratch under the same conditions. But the search time is less than an hour from a 5 GPU days pre-trained meta-network.

\end{abstract}

\section{Introduction}\label{introduction}
\label{intro}
%\cite{RefB} Your text comes here. Separate text sections with
Neural Architecture Search (NAS)~\cite{zoph2016neural} has seen significant progress on various computer vision tasks, such as image classification~\cite{real2019aging,cai2018proxylessnas}, object detection~\cite{ghiasi2019fpn,yao2021joint}, and semantic segmentation~\cite{liu2019auto,calisto2020adaen}.
NAS completes architecture design tasks by automating architecture engineering to alleviate the burden of manual architecture designing and heavy dependence on prior knowledge.
Despite their remarkable achievements, NAS requires expensive computing resources to search for the optimal neural architecture of the target task.

One line of existing NAS methods focuses on improving search efficiency to explore the large search space, reducing the search time from thousands of GPU days~\cite{zoph2016neural,zoph2018learning} to a few GPU days~\cite{cai2018proxylessnas,liu2018darts,pham2018efficient, cai2019once}. ENAS~\cite{pham2018efficient} trains a supernet to share the parameters among subnets, thus reducing the search time to several GPU days.
Similarly, more and more NAS methods~\cite{yang2020cars,guo2020single,zhang2020one} optimize model weights by training a super network with shared parameters and optimize architecture by the evolutionary algorithm. These approaches are summarized into a more systematic framework named one-shot architecture search.
In addition, DARTS~\cite{liu2018darts} continuously relaxes the discrete architecture with all candidate operations, which makes it possible to jointly optimize architecture parameters and weight parameters by gradient descent.

However, most NAS approaches are only designed for specific tasks, since large scale and diverse datasets (mainly labeled data) are available for these tasks.
This makes it difficult to apply NAS in a multiple-task scenario since the dataset is a collection of small tasks with a few samples.
To meet such a need, learning algorithms should be adapted to new tasks rapidly by extracting learning experiences from existing related tasks, namely, meta-learning.
Model-agnostic meta-learning(MAML)~\cite{finn2017model} estimates a good meta initialization of model parameters for the fast adaptation to new tasks purely by a gradient-based search. Although MAML and its variants~\cite{antoniou2018train,finn2017model,nichol2018first} are applied successfully to a variety of learning tasks, existing algorithms mainly optimize the weights while ignoring the architecture, which is more macro in a network.

%Meta architecture search aims at learning task-agnostic knowledge that can be used to search over multiple tasks efficiently.
%Comparatively speaking, there has been some recent works on meta architecture
Meta architecture search learning task-agnostic knowledge that is used to search networks across multiple tasks efficiently.
Comparatively speaking, there has been some recent works on meta architecture search~\cite{shaw2019meta,kim2018auto,elsken2020meta,lian2019towards} by integrating meta-learning with NAS. Bayesian Meta Architecture Search~\cite{shaw2019meta} uses a Bayesian formulation of the architecture search problem to learn over an entire set of tasks simultaneously. Lian et al.~\cite{kim2018auto, lian2019towards} roughly combined MAML and DARTS to find the task-sensitive architecture by updating architecture parameters and weight parameters at the same time. Furthermore, Elsken et al.~\cite{elsken2020meta} proposed joint meta-learning of architecture and weights with METANAS. These approaches all find the task-sensitive architecture by updating architecture parameters and weight parameters at the same time. To search for the best meta-architecture, we try to combine various DARTS~\cite{liu2018darts} algorithms with MAML~\cite{finn2017model}. But as shown in Table~\ref{tab:table1}, existing DARTS-type algorithms~\cite{liu2018darts,chen2019progressive,xu2019pc} require a lot of GPU memory in training time. Simply training all candidate architecture leads to GPU memory explosion, as the memory consumption grows linearly with respect to the candidate number. In other words, GPU memory limits the search space of DARTS smaller than that of EA-based NAS~\cite{yang2020cars,guo2020single,lu2019nsga}. The flaw of DARTS eliminates the possibility of searching deeper networks with better performance.

To solve this issue, we present our effort towards meta architecture search based on meta-learning and EA-based NAS (less GPU memory), namely across task neural architecture search(AT-NAS). A continuous meta-evolution strategy is developed to maximally utilize the meta-knowledge that is learned from multiple tasks in the last evolution generation. Specifically, considerable cells are used to initialize a SuperNet with all candidate architectures.
Thus, only one candidate operation is selected at run-time, which reduced the required memory to the same level as running a compact model. Individuals of the architecture population are generated through several benchmark operations(i.e., crossover, mutation, and random sample). According to the mean accuracy of multi-task, architectures with high accuracy are preserved and corresponding cells in the SuperNet will be updated for subsequent optimization. In addition, the proposed distributed parallel meta-search and performance predictor~\cite{lu2020nsganetv2,lu2021neural,xu2021renas} aim to accelerate the search process of AT-NAS.

AT-NAS exploits the similarities among images of tasks and the corresponding optimal networks to learn a meta-architecture that can quickly adapt to new tasks in less time.
The superiority of AT-NAS is verified on benchmark datasets over previous meta architecture search methods.
The contributions of the paper are summarized as follows:
\begin{itemize}
\item A new approach AT-NAS is proposed to automaticly search of the meta-architecture by meta training the supernet and meta evolution algorithm. The optimization of meta-architecture for the specific task can be conducted with a little labled data and only a few iteration steps(less time consumption and higher accuracy).
\item Since the decoupling of architecture and parameter optimization, the search process of AT-NAS is not significantly limited by GPU memory resources. In addition, We propose the distributed parallel meta-learning and online performance predictor to further accelerate the search.
\item Extensive experiment results over multiple datasets show that the architecture of a new task by AT-NAS achieves better performance than search from scratch, but with 60$\times $ less search cost.
\end{itemize}

%The overall graphical illustration of the model can be found in Figure~\ref{fig:atnas}

%\begin{figure}[htbp]
%\begin{center}
%\includegraphics[width=9cm]{ATNASill.eps}
%\end{center}
%\caption{The main illustration of the proposed Across Task Neural Architecture Search, .}
%\label{fig:atnas}
%\end{figure}

\begin{table}
\centering
\caption{GPU memory-usage of various NAS methods. Initial network settings: 16 initial channels, 10 classifications, 8 layers and a random tensor $\left[ 1,3,224,224 \right] $ as input data}
\label{tab:table1}
\begin{tabular}{l|c}
\hline
Methods  & Memory-usage(M) \\ %& Flops(G) \\
\hline
DARTS~\cite{liu2018darts}        & 10050  \\ %& 15.64   \\
\hline
P-DARTS~\cite{chen2019progressive}      & 6611  \\ %& 9.74      \\
\hline
PC-DARTS~\cite{xu2019pc}      & 3235  \\%& 3.05    \\
\hline
\textbf{AT-NAS(ours)}  & \textbf{2044 $\pm$ 272}  \\ %& \textbf{2.74 $\pm$ 0.8}     \\
\hline
\end{tabular}
\end{table}

%\cite{breiman2001random}
%\section{Related Works}\label{relatedwork}
%\label{sec:1}
%Text with citations \cite{RefB} and \cite{RefJ}.
\section{Preliminary and Problem Formulation}\label{problem}
In this section, we formulate the meta architecture search problem. The whole dataset is divided into meta-train dataset $\mathcal{D} _{meta-train}$ and meta-test dataset $\mathcal{D} _{meta-test}$.
Consider a distribution over relatively small training tasks $\mathcal{T}^{train}\sim p^{train}\left( \mathcal{T} \right) $ in $\mathcal{D} _{meta-train}$ and test tasks $\mathcal{T}^{test}\sim p^{test}\left( \mathcal{T} \right) $ in $\mathcal{D} _{meta-test}$. Same as the task setting of few-shot learning~\cite{vinyals2016matching}, $n$-way, $k$-shot tasks entail that each task is a small classification problem with $n$ classes and $k$ training examples per class.

The meta-architecture $\mathcal{A}_{meta}$ with corresponding model weights $\mathcal{W}_{meta}$ can efficiently adapt to a new task $\mathcal{T}_{i}$ by learning a few labeled samples. In line with MAML~\cite{finn2017model}, we do so by minimizing a meta-objective function, which can be formulated as:

\begin{equation}\label{eq1}
\begin{split}
&\mathcal{L} _{meta}\left( \mathcal{W} ,\mathcal{A} ,p^{train},f \right)\\
&=\sum_{\mathcal{T} _i\sim p^{train}}{\mathcal{L} _{\mathcal{T} _i}\left( f^k\left( \mathcal{W} ,\mathcal{A} ,\mathcal{D} _{support}^{\mathcal{T} _i} \right) ,\mathcal{D} _{query}^{\mathcal{T} _i} \right)}\\
&=\sum_{\mathcal{T} _i\sim p^{train}}{\mathcal{L} _{\mathcal{T} _i}\left( \left( \mathcal{W} ',\mathcal{A} ' \right) ,\mathcal{D} _{query}^{\mathcal{T} _i} \right)}
\end{split}
\end{equation}
$\mathcal{T} _{i\,\,}=\left( \mathcal{D} _{support}^{\mathcal{T} _i},\mathcal{D} _{query}^{\mathcal{T} _i} \right)$ denotes a training task, $\mathcal{L} _{\mathcal{T} _i}$ is the corresponding task loss and $f^k\left( \mathcal{W} ,\mathcal{A} ,\mathcal{T} ^{train} \right)$ represents the  task learning algorithm or simply task-learner, where $k$ refers to the iteration number of learning.
$f^k\left( \mathcal{W} ,\mathcal{A}_{fixed} ,\mathcal{T} ^{train} \right)$ (where architecture is predefined and fixed) only performs $k$ update to weights within each task. Then, the shared weight $\mathcal{W}$(meta weight) can be updated by summarizing the task-specific parameters and an optimizer like SGD. The task-learner is optimized according to the following rule:

\begin{equation}\label{eq2}
\begin{split}
\mathcal{W} _{i}^{m+1}=\mathcal{W} _{i}^{m}-\lambda _{inner}\nabla _{\mathcal{W} _{i}^{m}}\mathcal{L} _{\mathcal{T} _i}\left( \mathcal{W} _{i}^{m},\mathcal{D} _{support}^{\mathcal{T} _i} \right)
\end{split}
\end{equation}
where $\mathcal{W} _{i}^{0}=\mathcal{W}$ and $\lambda _{inner}$ is a task-inner learning rate. The meta-learner is optimized according to the following rule:
\begin{equation}\label{eq3}
\begin{split}
\mathcal{W} =\mathcal{W} -\lambda _{meta}\nabla _{\mathcal{W}}\mathcal{L} _{meta}\left( \mathcal{W} ,\mathcal{D} _{query} \right)
\end{split}
\end{equation}
where $\mathcal{D} _{query}$ includes several task query sets (a meta mini-batch of tasks), and $\lambda _{meta}$ is a meta learning rate. Meta-learning of architecture parameters $\mathcal{A}$ is directly implemented by using a gradient-based NAS algorithm as a task optimizer $f^k$, like DARTS

However, such a crude combination causes GPU memory explosion if searching a deep network. Refer to the existing methods~\cite{liu2018darts,hu2020dsnas,xie2018snas}, the core of gradient-based NAS is to continuously relax the discrete architecture with all possible operations and jointly optimize the architecture parameters and network weights by gradient descent. In other words, architecture parameters and network weights consume more GPU memory than that of a fixed-architecture network. In addition, architecture parameters and weights naturally have a relationship of competition and cooperation in the DARTS-type strategy, which leads to a poor performance in long-term training, as too many skip-connection in searched architecture. Therefore, we are committed to address the challenge that training meta-parameters and search a meta-architecture during the process with stable GPU memory occupation.

\section{Across Task Neural Architecture Search}\label{AT-NAS}
\begin{figure*}[htbp]
\begin{center}
\includegraphics[width=11cm]{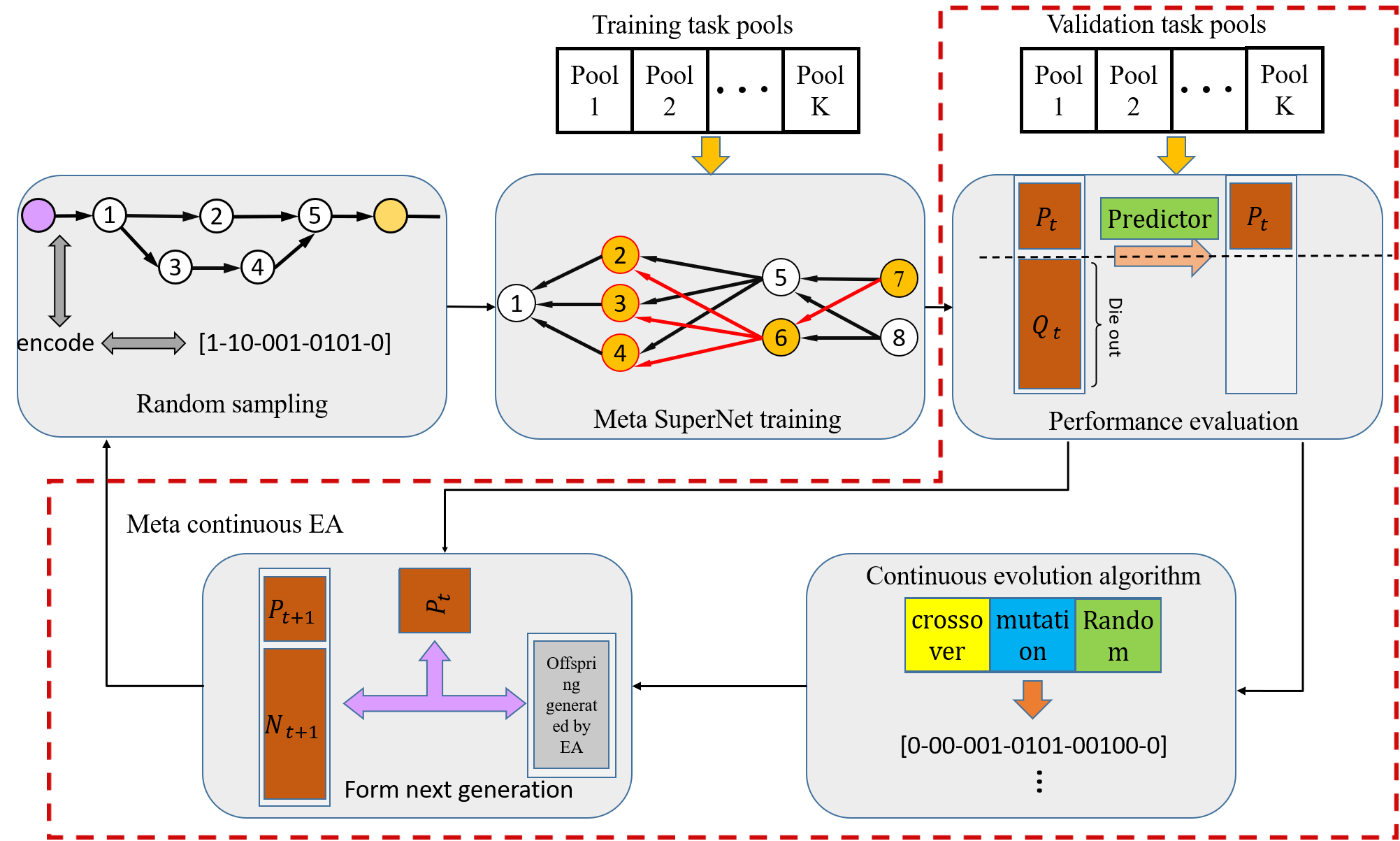}
\end{center}
\caption{The main framework of the proposed Across Task Neural Architecture Search strategy.}
\label{fig:AT-NAS}
\end{figure*}

In this section, a novel across-task neural architecture search(AT-NAS) is designed, which includes meta parameter optimization and meta architecture optimization.
%A SuperNet $\mathcal{N}$  shares parameters $\mathcal{W}$ for different architectures and use the genetic algorithm for generating a set of well-performed architectures that cover a vast space.
A SuperNet $\mathcal{N}$ shares parameters $\mathcal{W}$ for different architectures and generates a set of well-performed subnets that cover a vast space by genetic algorithms.
After that, distributed parallel meta-search and the performance predictor is designed to accelerate the search process of AT-NAS. Finally, an instance is presented to indicate that the searched meta-architecture adapts to a new task by a few gradient descent steps. The overall framework of the AT-NAS can be found in Figure~\ref{fig:AT-NAS}, where the main flow path is the EA-based NAS fully integrating gradient-based meta learning.

\subsection{SuperNet of AT-NAS}\label{subs1}
AT-NAS starts with optimizing the SuperNet via MAML-like meta training on multiple tasks. The generality of the selected search space has a major impact on the quality of feasible results. AT-NAS is evaluated on a popular design of the overall structure of the SuperNet (same as DARTS~\cite{liu2018darts}) to compare with previous approaches and illustrate the effectiveness. The SuperNet backbone is conducted by normal cells and reduction cells. The normal cell is used for the layers with the same spatial size of input and output features. The reduction cell is used for layers that downsample the input features.

The cell comprises two input nodes, four intermediate nodes, one output node, and fourteen edges. The two input nodes are the output of the previous two cells. The intermediate node is the sum of the output features of previous nodes after passing through edges. The output node combines the output features of each intermediate node. Edges represent candidate operations (a total of eight operations: separable convolutions $3\times3$ and $5\times5$; dilated separable convolutions $3\times3$ and $5\times5$; max pooling $3\times3$, average pooling $3\times3$; identity; none-operation). The ReLU-Conv-BN structure is used for convolutional operations and applied twice in each separable convolution. The network is stacked with a certain number of searched cells, general 2 or 8 cells.

\begin{figure}[htbp]
\centering

\subfigure[Illustration of DARTS search space] {
    \begin{minipage}[t]{0.48\textwidth}
        \centering
        \includegraphics[width=6cm]{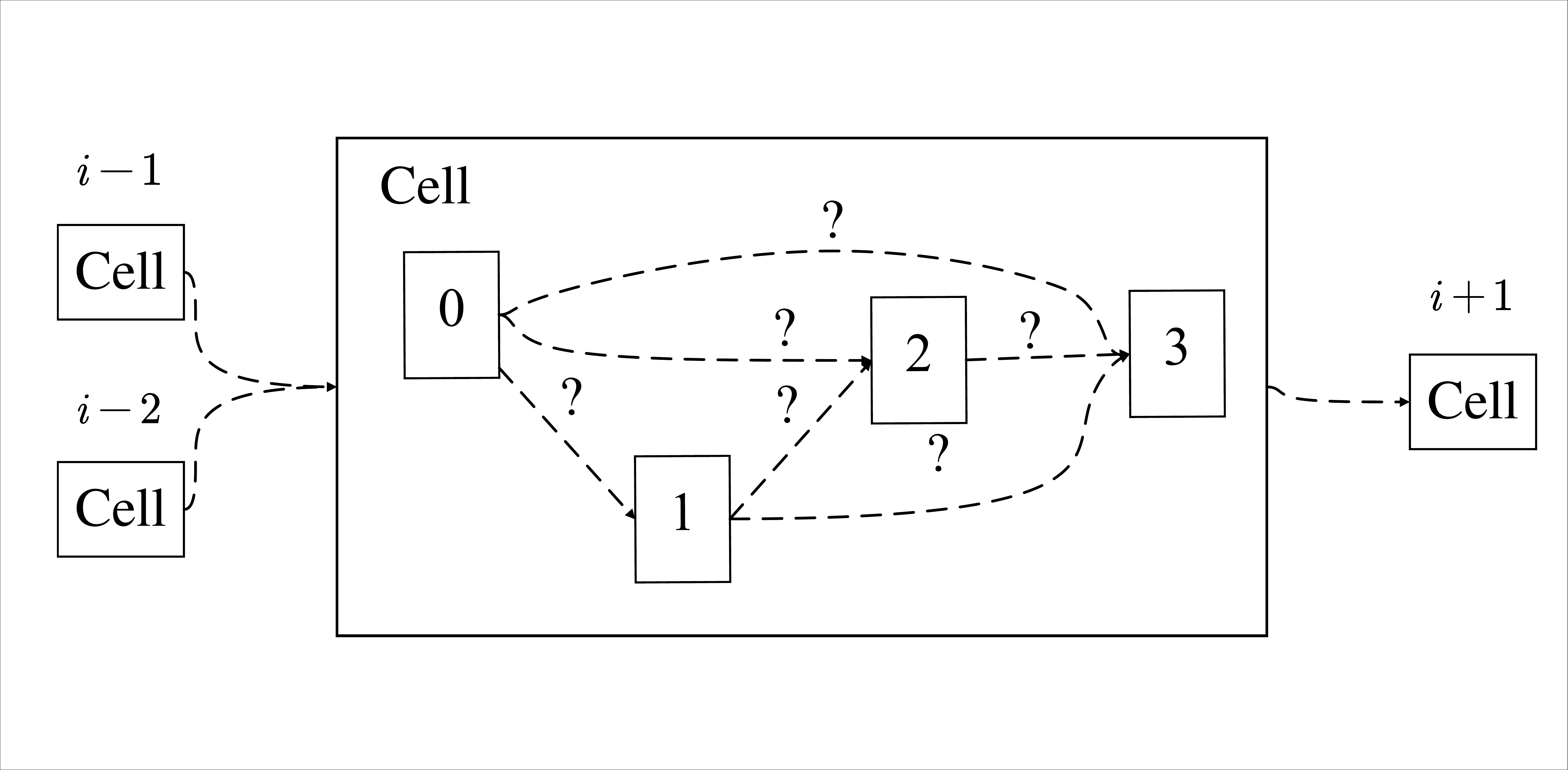}
        \label{fig:a}
    \end{minipage}
}
\subfigure[Architecture code] {
    \begin{minipage}[t]{0.48\textwidth}
        \centering
        \includegraphics[width=6cm]{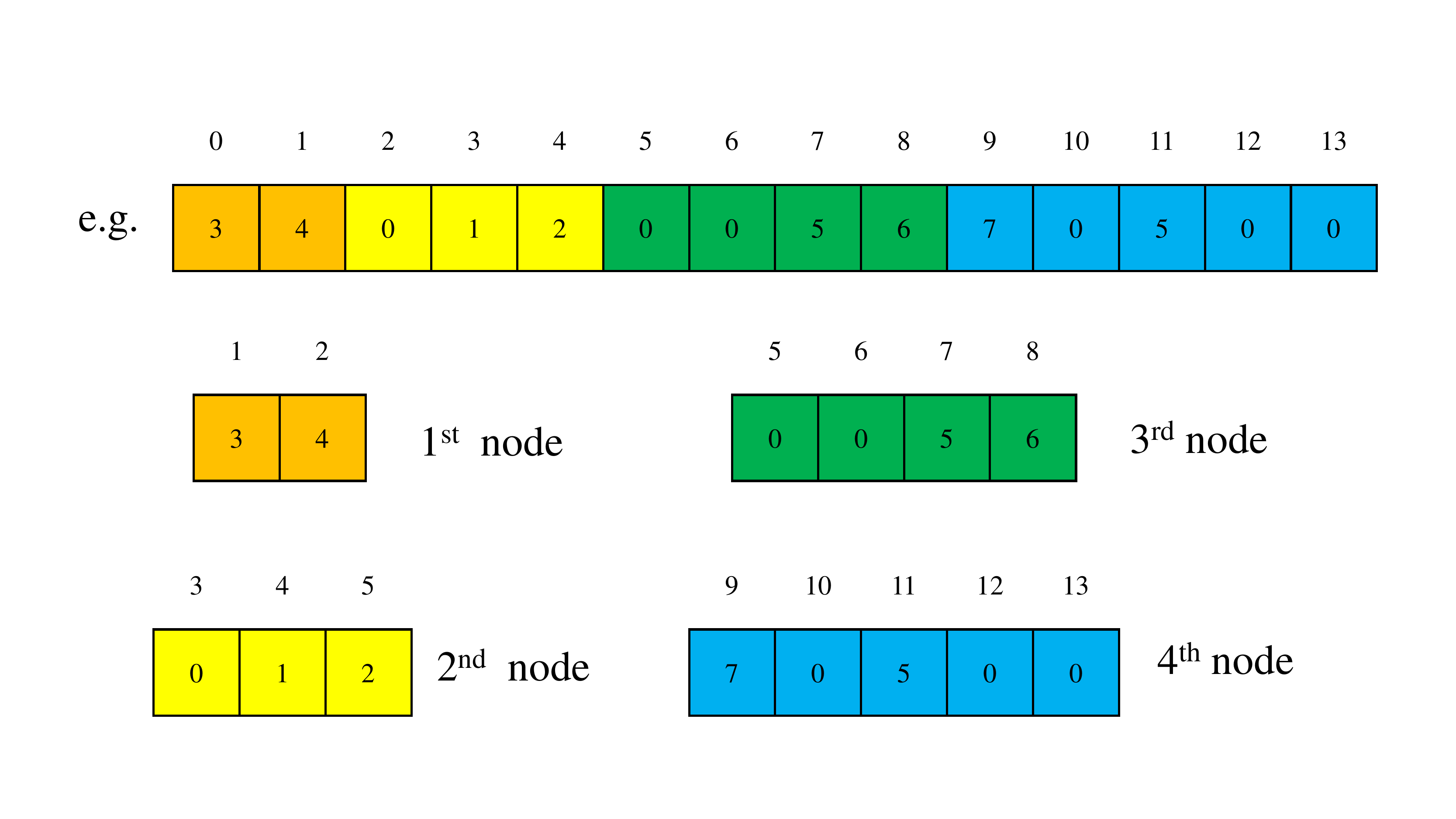}
        \label{fig:b}
    \end{minipage}
}
\caption{(a): The architectures in AT-NAS are variants of DARTS (b): the architecture code of cell is composed of bits that are corresponding to operations.}
\label{fig:DARTS}
\end{figure}

Each subnet ${N}_{i}$ is represented by a set of $\mathcal{W}_{i}$ randomly sampled from the SuperNet $\mathcal{N}$ and an architecture sequence $\mathcal{A}_{i}$. The architecture sequence is an integer string of length $14\times c$, where $c$ is the number of cells. The value of each bit expresses the possible operations in the corresponding edge. Observing the network searched by DARTS, we design a restriction of sampling subnets:
each intermediate node must have two edges with certain operations and the first intermediate node only has two edges with certain operations. Therefore, only two bits of each node have the value of operations and others are zero, as shown in Figure~\ref{fig:DARTS}.

%Aiming at large-scale dataset transfer learning and running on the mobile platform, how to ensure that as much GPU memory as possible is used for larger learning tasks is the main problem. However, the multi-path of previous search space occupy more GPU memory, which seriously hinder the improvement of search efficiency. Therefore, the choice block of SPOS~\cite{guo2020single} is considered as another stage of the SuperNet. We refer to the above two search spaces as \textbf{S1} and \textbf{S2} in the following content.

%补超网示意图
%Note that the sampling and representation strategies of architectures are the same in (\text{S2}) and(\text{S2}).

From the above, it is seen that each subnet $\mathcal{N}_i$ could be represented as a $\left( \mathcal{W}_i,\mathcal{A}_i \right) $ pair. The SuperNet parameters $\mathcal{W}$ are shared by a set of subnet parameters $\mathcal{W}_i$. If the shared parameters of SuperNet and the architecture population are both randomly initialized, the most frequently used operations for all the architecture would be trained more times. Following one-shot NAS methods, candidate architectures follow the same probability of being sampled to participate in training for several epochs, which is called warm-up. The parameters can be optimized by MAML or REPTILE~\cite{nichol2018first}, while architectures $\mathcal{A}$ are alternately optimized by the meta continuous evolution algorithm. Finally, the optimal $\mathcal{W}$ will fit all searched subnets for achieving higher mean recognition accuracy on multiple tasks.
The above two steps constitute the main optimization part of our proposed AT-NAS pipeline. We will introduce these two optimization steps in the following.
%Finally, the optimal $\mathcal{W}$ will fit for all subnets to achieve higher mean recognition accuracy on multiple tasks.

\subsection{Meta Parameter Optimization}\label{subs2}
The parameters $\mathcal{W}$ of the supernet are seen as the collection of all the architecture parameters. The parameters $\mathcal{W}_{i}$ of the $i$-th individual are $\mathcal{W} _i=\mathcal{W} \odot \mathcal{A} _i$, where $\odot$ is the selection operation that keeps the parameters of complete feature maps only for the operations corresponding to the position in the sampled architecture.

Denote $X$ as the input data of a task $\mathcal{T}_{j}^{train}$, the prediction result of the current subnet is $\mathcal{N}_i\left( X \right) $, where $\mathcal{N}_i$ is the $i$-th subnet architecture. The loss of the $i$-th subnet on the $j$-th task can be expressed as $\mathcal{L}_{i,j}=\mathcal{H}_{\mathcal{T}_j}\left( \mathcal{N}_i\left( X \right) ,Y \right) $, $\mathcal{H}_{\mathcal{T}_j}$ is the evaluation criterion of the $\mathcal{T}_j$ task. On the support set of $j$-th task, the $i$-th subnet is optimized according to the following formula:

\begin{equation}\label{eq4}
\begin{split}
\mathcal{W} _{i,j}^{m+1} &= \mathcal{W} _{i,j}^{m}-\lambda _{task}\nabla _{\mathcal{W} _{i,j}^{m}}\mathcal{L} _{i,j}\\
&=\mathcal{W} _{j}^{m}\odot \mathcal{A} _i-\lambda _{task}\nabla _{\mathcal{W} _{j}^{m}\odot \mathcal{A} _i}\mathcal{L} _{i,j}
\end{split}
\end{equation}

In practice, calculating high-order derivatives is computation inefficient. The finite difference approximation of the gradient is commonly used to replace the calculating second derivatives in MAML.
This is similar to the approximate processing of First-order MAML~\cite{antoniou2018train} and REPTILE~\cite{nichol2018first}, which can be formulated as

\begin{equation}\label{eq5}
\begin{split}
\mathcal{W} ^{n+1}=&\mathcal{W} ^n-\eta _{meta}\frac{1}{K\cdot B_{\mathcal{T} _{q}^{train}}}\cdot
\\
&\sum_{i=1}^K{\sum_{j=1}^{B_{\mathcal{T} _{q}^{train}}}{\nabla _{\mathcal{W} _{i,j}}\mathcal{L} _{i,j}\left( \mathcal{T} _{q}^{train} \right)}}
\end{split}
\end{equation}
where $K$ is the number of sampled subnets and $\eta _{meta} $ is meta learning rate. $B_{\mathcal{T} _{q}^{train}}$ is the number of a batch of training tasks. $\nabla _{\mathcal{W} _{i,j}}\mathcal{L} _{i,j}\left( \mathcal{T} _{q}^{train} \right)$ represents the gradients of meta parameters on the query sets of training tasks. The parameters are only optimized by networks that use corresponding operations during forwarding. By collecting the individual gradients in the population, the parameters $\mathcal{W}$ are updated by the SGD algorithm.

Training time is greatly extended as there is a large set of architectures with shared weights of SuperNet. Accumulating the gradient of all architectures would take much time for one-step gradient descent. Borrowing the idea of SGD, we randomly extract mini-batch architectures from the population to update parameters. $B_{arch}$ different architectures where $B_{arch}<K$, and the indices of architectures are $\left\{ n_1,\dots ,n_{B_{arch}} \right\}$. The efficient parameter optimization rule in Eq.~\ref{eq5} can be written as:
\begin{equation}\label{eq6}
\begin{split}
\mathcal{W} ^{n+1}\approx \mathcal{W} ^n-\eta _{meta}\frac{1}{B_{arch}\cdot B_{\mathcal{T} _{q}^{train}}}\cdot
\\
\sum_{i=1}^{B_{arch}}{\sum_{j=1}^{B_{\mathcal{T} _{q}^{train}}}{\nabla _{\mathcal{W} _{i,j}}\mathcal{L} _{i,j}\left( \mathcal{T} _{q}^{train} \right)}}
\end{split}
\end{equation}
On a meta mini-batch of tasks, the gradients over a mini-batch of architectures are seen as an unbiased approximation of the averaged gradients of the $K$ individuals. The time cost is largely reduced and architecture mini-batch size $B_{arch}$ is seen as a hyper-parameter to balance efficiency and accuracy, same as meta mini-batch size.
Moreover, we can further accelerate learning via using distributed parallel computation, specifically demonstrated in Section~\ref{subs4}.

\subsection{Meta Architecture Optimization}\label{subs3}
%The fundamental idea of MAML is to accumulate the experience (gradient) acquired on multiple tasks to update the meta-network.
Inspired from MAML, a meta continuous evolution algorithm is proposed to accumulate the experience (recognition performance) on multiple tasks. A population with quick task adaptation is maintained via the proposed meta continuous evolutionary algorithm. The impact of the generation of individuals is second only to the SuperNet. Regardless of the search space, the genes of individuals in the population are composed of architecture codes.

There is a one-to-one correspondence between the values of architecture codes and candidate operations, except for "0" as no operation. Preset rules restrict the generation of the architecture code. These rules are related to other evaluation metrics rather than accuracy, such as parameters number, float operations, latency, energy. In this paper, float operations are generally used as additional constraints. We also consider the case of the final compact model. For example, two operations are usually preserved between nodes of the compact model. Therefore, the encoding position of the corresponding node is defined to have only two non-zero values, and the others are all zero. This strategy can ensure accelerated convergence and facilitate comparison with previous methods.

Two pools of tasks $\mathcal{T}^{train}$ and $\mathcal{T}^{val}$ are constructed according to the setting of few-shot learning. SuperNet is initialized as described in Section~\ref{subs1}. The population $\left\{ \mathcal{A} _1,\dots \mathcal{A} _K \right\}$ is maintained by $K$ architectures, where $K$ is a hyper-parameter. Given the SuperNet $\mathcal{N}$, a batch of tasks from the training tasks pool $\mathcal{T}^{train}$ and a population of architecture $\mathcal\bf{A}$, the proposed meta-parameter optimizing scheme is used for updating supernet parameters. After several iterations, we evaluate the rapid learning ability of individuals in the population on several batches of tasks. The corresponding subnet is updated on the support set of each task following Eq.~\ref{eq4}.
The task-inner training iteration period only contains $M$ iterations to filter out the subnets with fast convergence, where $M$ is set to a relatively small integer. Then we test the trained subnet on the query set of the corresponding task to get the accuracy. The average accuracy of the single subnet learning multiple tasks is used as the fitness value of this subnet. We design the pseudo-code of the concrete algorithm for Across-Task Neural Architecture Search (AT-NAS) in Algorithm ~\ref{alg:at-nas_algorithm}.

Evolution algorithm is used to complete the architecture search and Algorithm~\ref{alg:EA} for the pseudo-code according to the fitness. During the evaluation step, $t\times K$ individuals with high fitness are directly used as direct offspring, where $t$ is the hyper-parameters controlling the generation of offspring, generally set to $1/4$. Considering the difference in search spaces, we design appropriate crossover and mutation to produce other offspring.

\begin{figure}[htbp]
\begin{center}
\includegraphics[width=8cm]{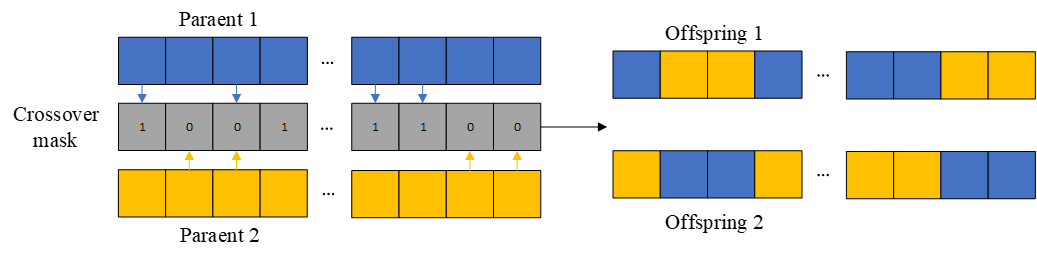}
\end{center}
\caption{The offspring architectures are generated by recombining codes of the two-parent architectures. The probability of any parent selected is equal. The number of offspring is the same as the parent}
\label{fig:crossover}
\end{figure}

\textbf{Crossover} allows two population members to exchange information for generating two new members. A customized homogeneous crossover~\cite{lu2021neural} is adapted to uniformly generate offspring architectures from parent architectures. Regardless of search space difference, we visualize the implementation of the crossover operation in Figure~\ref{fig:crossover}. The crossover operation generates an offspring population of the same size as the parent population in each generation. In addition, code positions of the corresponding node in the offspring code are also designed to have only two non-zero values. This ensures that the nature of the offspring is consistent with that of the parent.

\textbf{Mutation} is a local operation that introduces randomness into the population to get the desired result faster. There is a tiny probability $p_m$ that a certain part of the gene code of the retained individual may mutate into other situations when copying. It may be the same as the original individual after the copy is completed, or some of the gene codes may be different. The mutation of the code value is random, but like crossover, we restrict the mutation operation in a predefined way.

The mutation and crossover offspring population are then merged with the parent population following Algorithm ~\ref{alg:EA}. Significantly, if the population size is smaller than $K$ after crossover and mutation, the vacant part of the population is randomly sampled from the SuperNet. Following this way, the population size remains at $K$ until the end of evolution. The final population is sensitive to new tasks and can quickly converge to the optimal network via EA-based NAS on a new task. As described in ~\ref{subs5}, the evolutionary search based on the meta continuous evolutionary algorithm is embedded in the meta architecture optimization, which iteratively runs together with the meta parameter optimization.

\begin{algorithm}[tb]
\caption{Across Task NAS via Meta-Learning}
\label{alg:at-nas_algorithm}

\textbf{Input}:Distribution over tasks $p\left( \mathcal{T} \right) $, Task loss fuction $\mathcal{L}$, Supernet $\mathcal{N}$\\
\textbf{Parameter}:Step size hyperparameters $\lambda _{task}$, $\epsilon $ \\
\textbf{Output}:A meta-network population including the meta supernet $\mathcal{N}_{meta}$ and the meta-architecture population $\mathcal{A}_{meta}$
\begin{algorithmic}[1] %[1] enables line numbers
\STATE Initialize  supernet $\mathcal{N}$ with $\mathcal{W}_{meta}$ and $\mathcal{A}_{meta}$
\STATE Warm up the SuperNet $\mathcal{N}$ for $E_{warm}$ epochs
\FOR{$e=1,\dots ,E_{evo}$}
\STATE Sample meta task $\mathcal{T}_1\dots \mathcal{T}_n$ from $p\left( \mathcal{T} \right) $
\FOR {all $\mathcal{T}_j$}
\STATE Random sample architecture $\mathcal{A}_1\dots \mathcal{A}_B$ from $\mathcal{A}_{meta}$ and select the corresponding $P$ collections of weights $\mathcal{W}_1\dots \mathcal{W}_b$ from $\mathcal{N}$ to form the $B$ subnets $\mathcal{N}_1\dots \mathcal{N}_B$
\FOR{all $\mathcal{N}_i$}
\FOR {$m=1,\dots ,M$}
\STATE $\mathcal{W} _{i,j}\gets \mathcal{W} _j\odot \mathcal{A} _i-\lambda _{task}\nabla _{\mathcal{W} _j\odot \mathcal{A} _i}\mathcal{L} _{i,j}$
\ENDFOR
\ENDFOR
\STATE $\mathcal{W} \gets \mathcal{W} -\eta _{meta}\frac{1}{B_{arch}\cdot B_{\mathcal{T} _{q}^{train}}}\cdot\sum_{i=1}^{B_{arch}}
{\sum_{j=1}^{B_{\mathcal{T} _{q}^{train}}}{\nabla _{\mathcal{W} _{i,j}}\mathcal{L} _{i,j}\left( \mathcal{T} _{q}^{train} \right)}}$
\STATE Evaluate the trained subnets on the validation task pool and get the corresponding accuracy as the fitness of each subnet
\STATE Updating $\left\{ \mathcal{A}_1,\dots \mathcal{A}_P \right\} $ using evolutionary algorithm
\STATE Random sample subnets from supernet to supplement the population
\ENDFOR
\ENDFOR
\STATE \textbf{Return}  $\mathcal{N}_{meta}$ and $\mathcal{A}_{meta}$
\end{algorithmic}
\end{algorithm}

\begin{algorithm}[tb]
\caption{Evolutionary Search}
\label{alg:EA}
\textbf{Premise}:Evolutionary search is embedded in algorithm~\ref{alg:at-nas_algorithm} as the evolutionary algorithm \\
\textbf{Input}: Architecture population $\mathcal{A}$, SuperNet $\mathcal{N}$ with parameters $\mathcal{W}$, population size $K$, crossover probability $p_c$, mutation probability $p_m$\\
\begin{algorithmic}[1] %[1] enables line numbers
\STATE Compute the mean accuracy of each subnet on multiple tasks $f\left( \mathcal{A} _i,\mathcal{W} \right)$
\STATE Retain $top-K/4$ ranked individuals  as parents $P$
\STATE Create offspring by crossover between parents $Q_c\gets Crossover\left(P,p_c \right)$
\STATE Induce randomness to offspring through mutation $Q_m\gets Mutation\left(P,p_m \right)$
\STATE Rondom Sample architectures from SuperNet to supplement the population $Q_s\gets Sample(\mathcal{W})$
\STATE Merge parents and offspring $\mathcal{A}\gets P\cup Q_c\cup Q_m\cup Q_s$
\STATE \textbf{Return}  $\mathcal{A}$

\end{algorithmic}
\end{algorithm}

\subsection{Acceleration Strategy} \label{subs4}
%\paragraph{Search Time Analysis}
%During the search stage, train tasks are used for updating network parameters, and validation tasks are used for updating architectures. Assuming the average training time on one meta train task for one architecture is $t_tr$ and
\paragraph{Distributed Parallel Meta-search}
In common supervised learning, datasets and models are generally distributed to multiple GPUs for parallel computing to accelerate training neural networks. In our survey, MAML-related papers rarely explain how to run MAML algorithms in a distributed way. There are no cases of using distributed parallel computing in code works. When trying to parallelize MAML, we found that the meta-learning dataset setting is the reason that it is difficult to parallelize MAML. As introduced in Section~\ref{problem}, the dataset of meta-learning has a two-layer structure. A single classification task has only a few pictures. If the training data is minimal, the parallel calculation will cause the mean and variance of the batch normalization layer in the network of each process to be inaccurate. This leads to incorrect gradients of model weights in multiple processes and affects the final result. Moreover, the time consumed by parallel computation is much smaller than that consumed by system scheduling in the meta-learning process.

\begin{figure}[htbp]
\centering
\subfigure[Illustration of distributed parallel training] {
    \begin{minipage}[t]{0.48\textwidth}
        \centering
        \includegraphics[width=8cm]{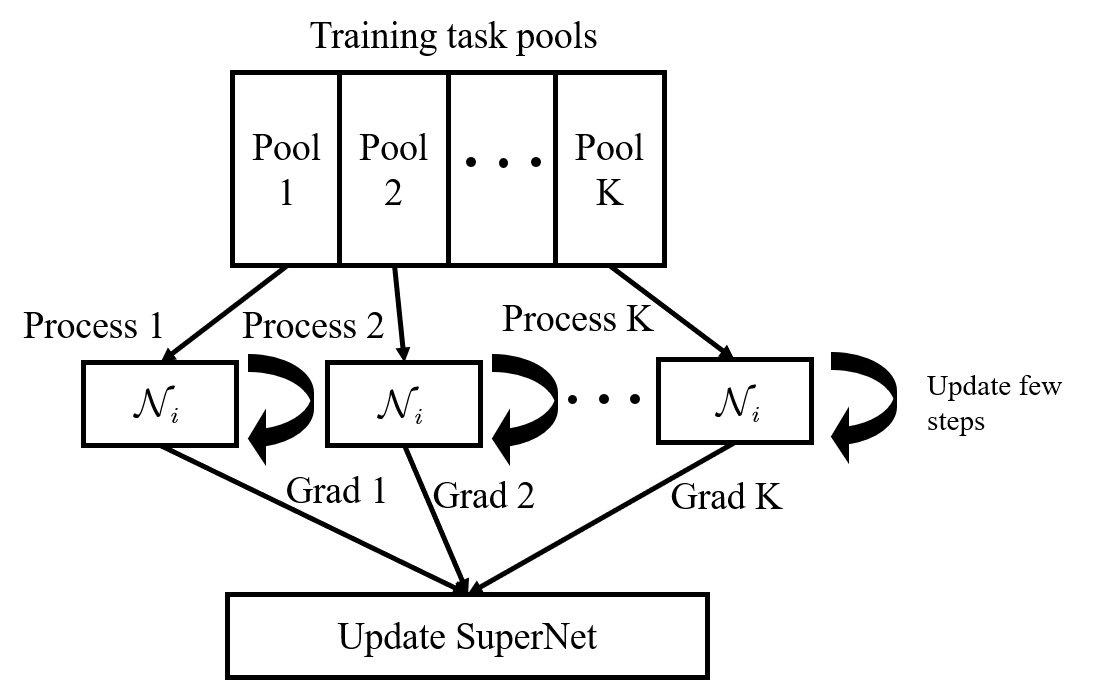}
    \end{minipage}
}
\subfigure[Illustration of distributed parallel valdation] {
    \begin{minipage}[t]{0.48\textwidth}
        \centering
        \includegraphics[width=8cm]{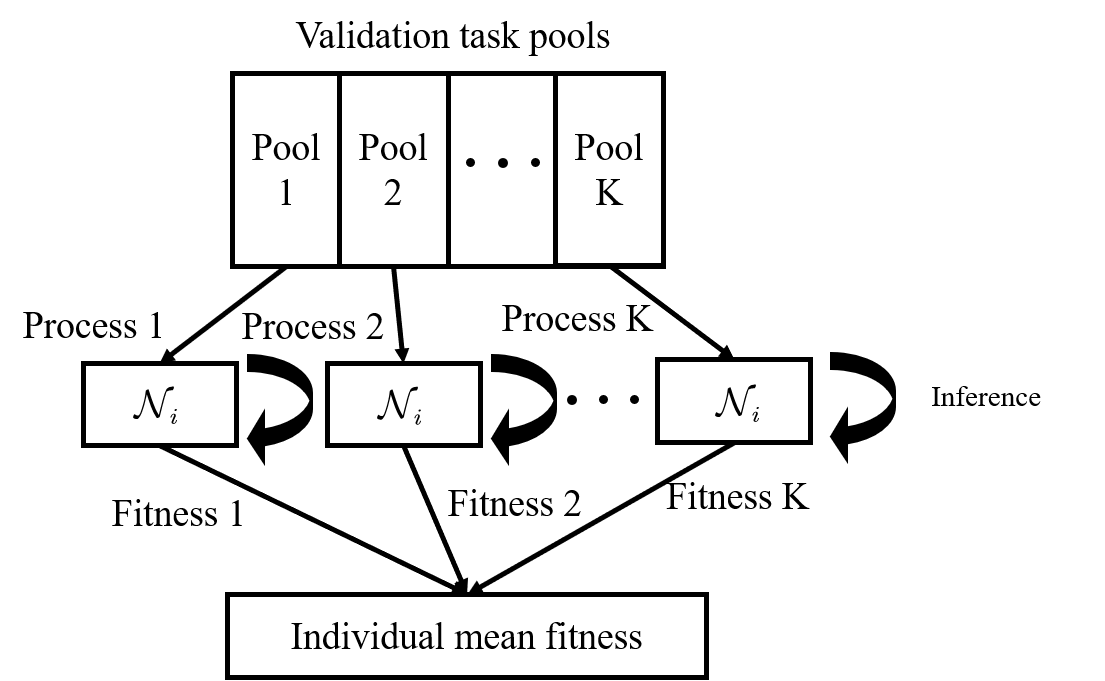}
    \end{minipage}
}
\caption{The task pool is equally divided into $K$ smaller task pools. The network in each process is trained independently. The average gradient and average fitness are used to update the SuperNet and evolutionary algorithms, respectively}
\label{fig:parallel}
\end{figure}

In response to this problem, we propose a new distributed parallel computing solution to accelerate meta architecture search. In practice, the training task pool is divided into $\mathcal{K} $ relatively small task pools in advance, where $\mathcal{K}$ is also the number of processes and available GPUs. Each process obtains the same network $\mathcal{N}_i$ and various training tasks in each iteration. The inner-task optimization process is executed according to Eq.~\ref{eq4} in each process. During the inner optimization, it is worth noting that the network parameters of each process are independent and do not exchange among processes.
When all processes have completed own inner optimization, the gradients of current training networks are accumulated to calculated the average gradients for meta parameters update. The specific parameters update is detailed as Eq.~\ref{eq7}
\begin{equation}\label{eq7}
\begin{split}
&\mathcal{W} ^{n+1}\approx \mathcal{W} ^n-\eta _{meta}\frac{1}{\mathcal{K} \cdot B_{arch}\cdot B_{\mathcal{T} _{q}^{train}}}\cdot
\\
&\sum_{k=1}^{\mathcal{K}}{\sum_{i=1}^{B_{arch}}{\sum_{j=1}^{B_{\mathcal{T} _{q}^{train}}}{\nabla _{\mathcal{W} _{i,j}}\mathcal{L} _{i,j}\left( \mathcal{T} _{q}^{train} \right)}}}
\end{split}
\end{equation}
means that the meta-network can complete the learning of $\mathcal{K}$ tasks at the same time. Finally, the collected average gradients are used as the meta-network update gradients in Eq.~\ref{eq6} to optimize the meta-network. The entire parallelized training and validation process is shown in Fig.~\ref{fig:parallel}.

The original time cost $t_{train}$ for the supernet to learn a complete task pool is reduced to $t_{train}/K$. Since the examples of each task will only be inferred by the network in the corresponding process (that is, a GPU), the variance and mean calculations for this batch of examples are correct and stable, avoiding the instability of the batch normalization layer. When running the evolutionary algorithm, we also divide the verification task pool evenly among $K$ processes. Finally, only the fitness value of the evaluated architecture in each process is averaged and used as the individual fitness of the architecture.

\paragraph{Performance Predictor}
The main computational bottleneck of NAS arises from the nesting caused by the bi-level optimization problem~\cite{lu2020nsganetv2}. Early NAS methods~\cite{zoph2016neural,zoph2018learning} require the subnet to be learned from scratch to evaluate the performance.
Weight-sharing strategies~\cite{pham2018efficient,cai2018proxylessnas,cai2019once} allow sampled subnets to inherit weights from the SuperNet, thereby effectively reducing the time consumption of learning for each subnet.
However, weight-sharing still requires the subnet inference on validation data to assess its performance (usually taking minutes).
Simply evaluating the subnets still requires a lot of time for the population consisting of thousands of architectures~\cite{tan2019mnasnet,real2019regularized}. To mitigate the time cost of a complete evaluation of subnets, we adapt a surrogate performance predictor to regress the performance of the sampled subnet without inference. Note that the predictor is only for a single task with a large number of samples. A task only contains dozens of pictures in few-shot learning so that the model inference time cost is not high. The performance of the predictor in actual few-shot learning experiments is relatively poor.

\begin{figure}[htbp]
\begin{center}
\includegraphics[width=8cm]{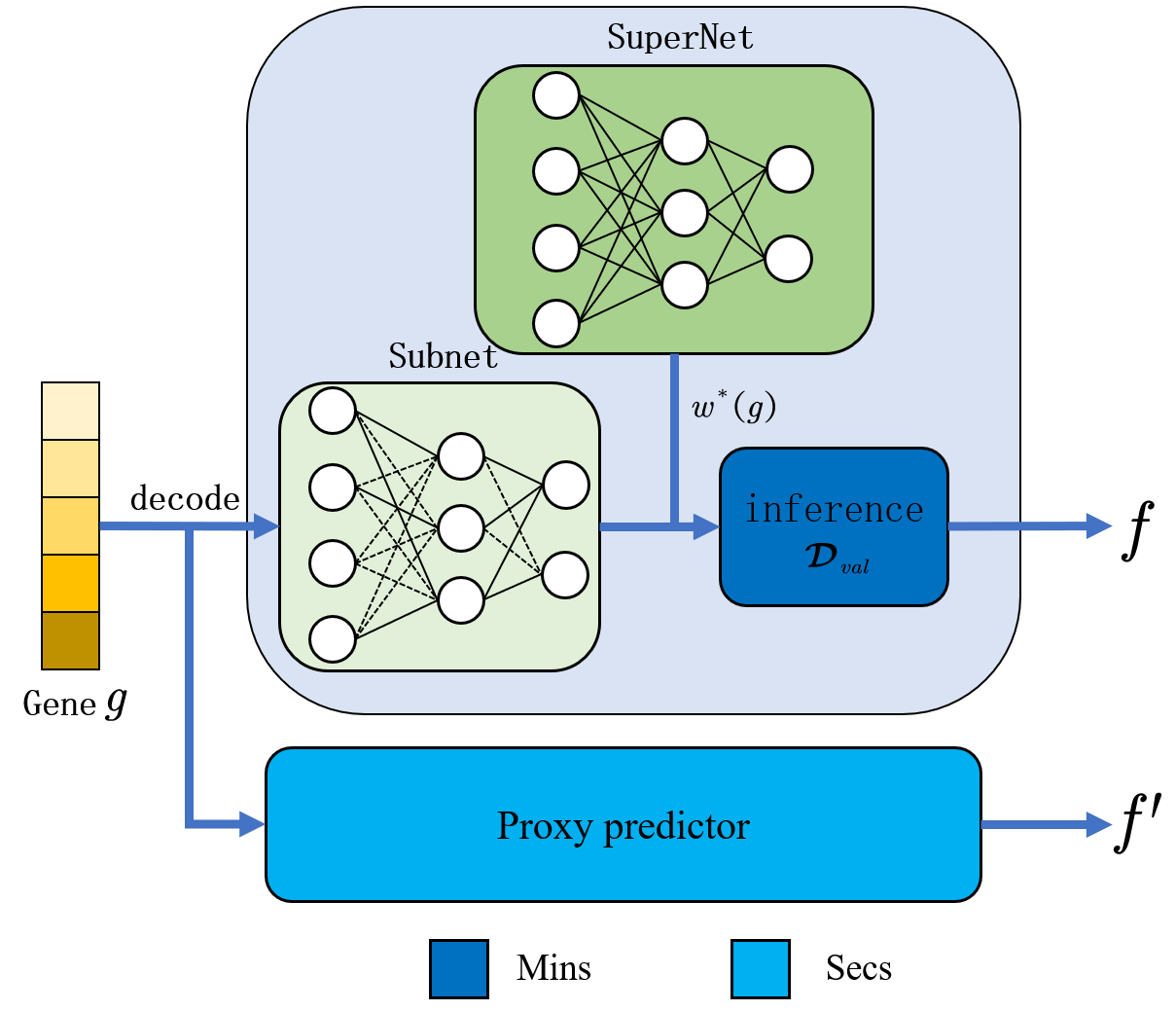}
\end{center}
\caption{Top path: A typical process of evaluating an subnet in weight-sharing NAS algorithm. Lower path: The performance predictor directly regresses the accuracy of the subnet from the architecture code (gene)}
\label{fig:predictor}
\end{figure}

The essence of the performance predictor is to obtain a function to represent the relationship between the integer-string (subnet code) and the corresponding performance by learning on a small amount of data, thereby greatly reducing the time for architecture evaluation.
This concept is illustrated in Fig.~\ref{fig:predictor}. The effectiveness of this method strictly depends on the quality of the proxy predictor and sample data.
We have defined several demands that a proxy predictor should have:

\begin{itemize}\label{itemize:demand}
  \item [1)]
  There is a high rank-order correlation between predictor and performance.
  \item [2)]
  The prediction performance should be stable across different datasets.
  \item [3)]
  The amount of data required to build a performance predictor is as small as possible.
\end{itemize}

\begin{figure}[htbp]
\begin{center}
\includegraphics[width=8cm]{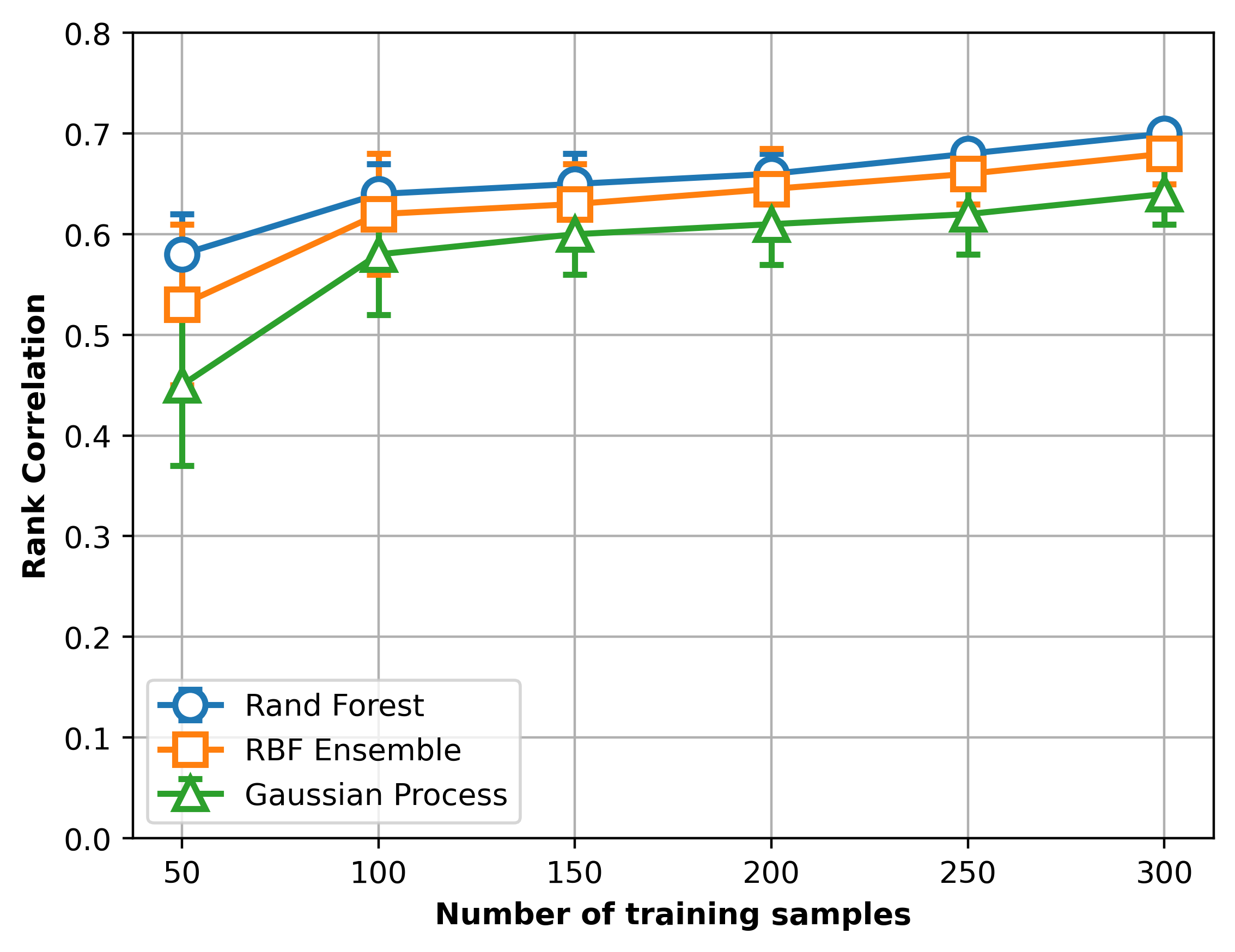}
\end{center}
\caption{ Accuracy predictor performance as a function of training samples. The size of Random Forest is 100}
\label{fig:correlation}
\end{figure}

By sampling $p$ samples from the current network population as the training data, the proxy predictor is limited to the search space, thereby ensuring a high-level correlation between prediction performance and performance.
To meet above demands, the average accuracy of the subnet served as the training sample is seen as the corresponding label.
We adapt three low-complexity online predictors, namely, Gaussian Process (GP)~\cite{dai2019chamnet}, Radial Basic Function~\cite{baker2017accelerating}, and Random Forest\cite{breiman2001random}. The mean and standard deviation of the Spearman rank correlation are calculated on 5 task datasets for each model. Figure ~\ref{fig:correlation} compares the mean Spearman rank correlation between the predicted result of each predictor and true accuracy(the training sample size is varied from 50 to 300). Empirically, Random Forest has a higher correlation compared to the other models. Similar conclusions are also verified in NAT~\cite{lu2021neural}. Quantitatively, predictor performance does not meet our expectations, as the accuracy gap between MAML and transfer learning.
The basic target of the performance predictor is to distinguish the pros and cons of architecture rather than predict the specific accuracy~\cite{xu2021renas}. Thus, we compared the predicted rank with the actual rank. The predictor can basically separate higher-performance architectures. Within the acceptable range of prediction accuracy, the performance predictor is adopted to accelerate the search process.

\subsection{Fast Architecture Transfer}\label{subs5}
This paper aims to learn a transferable architecture that can accomplish a new task through a small few learning steps.
The final product of AT-NAS is a set of meta networks, which is called meta-network population. The network of the meta-network population is composed of the architecture from the meta-architecture population and the corresponding weights from the meta-supernet.
Because the performance of a network with inherited weights is correlated with that of the same subnet fully trained from scratch. Intuitively, a better initial supernet and corresponding initial architectures can accelerate the convergence speed of architecture optimization and further improve search efficiency. The meta-network population is seen as a collection of subnets that can adapt to some new tasks efficiently.

On the basis of the meta-network population, the best task-specific architecture is screened out by Algorithm~\ref{alg:EA}.
However, the final result is still a collection of architecture. To improve convergence, we modify the continuous evolutionary algorithm as Algorithm~\ref{alg:transfer}. Not randomly sampling new individuals from the supernet, but new individuals are generated by crossover and mutation. Besides, the population is reduced by half after each evolution and finally shrinks to only an individual. This strategy is called population-halving.

\begin{algorithm}[tb]
\caption{Faster Architecture Transfer}
\label{alg:transfer}
\textbf{Input}: Meta-architecture population $\mathcal{A}_{meta}$, Meta SuperNet $\mathcal{N}_{meta}$ with parameters $\mathcal{W}$, criterion $\mathcal{H}$  \\
\textbf{Parameter}: Evolution number $E_{evo}$, Parameter optimization epoch $E_{evo}$
\begin{algorithmic}[1] %[1] enables line numbers
\FOR{$e=1,\dots ,E_{evo}$}
\FOR {$e=1,\dots ,E_{param}$}
\FOR {Mini-batch data $X$, target $Y$ in loader do }
%\STATE Select the $P$ collections of weights $\mathcal{W}_1\dots \mathcal{W}_P$ according to meta architecture population $\mathcal{A}_{meta}$ and
\STATE Generate the $P$ subnets $\mathcal{N}_1\dots \mathcal{N}_P$ from meta SuperNet $\mathcal{N}_{meta}$
\STATE Forward all subnets
\STATE Update the network parameters $\mathcal{W}$ based on gradient $d\mathcal{W}=\frac{1}{P}\sum_{i=1}^P{\frac{\partial \mathcal{L}_i}{\partial \mathcal{W}}\odot \mathcal{A}_i}$
\ENDFOR
\ENDFOR
\STATE Evaluating trained $\left\{ \mathcal{N}_1,\dots \mathcal{N}_P \right\} $ on validation set
\STATE Keep the top $1/4$ individuals in accordance with fitness value
\STATE Generate new $1/4$ individuals by crossover
\STATE Generate new $1/4$ individuals by mutation
\STATE Generate new $1/4$ individuals by random sample from SuperNet
\ENDFOR
\STATE \textbf{Return}  Optimal network $\mathcal{N}$
\end{algorithmic}
\end{algorithm}

\section{Experiments}

The effectiveness of AT-NAS is evaluated on the standard few-shot learning and supervised learning image classification benchmarks. The entire experimental process includes architecture search and architecture evaluation. The searched meta-network population is composed of a meta SuperNet and a meta-architecture population. In the architecture evaluation stage, the optimal meta-network is further finetuned by transferring it to multiple tasks. Fast architecture transfer is added to find the optimal meta-network when facing transfer learning tasks on large-scale datasets.

\subsection{Datasets}

For the standard few-shot image recognition benchmarks, Mini-ImageNet and Omniglot are widely used as a benchmark.

\textbf{Omnight}~\cite{lake2015human} consist of 1623 characters from 50 alphabets. 1200 characters are randomly selected for meta training. Each character contains 20 examples drawn by different individuals and the images are downsampled to $28 \times 28$.

\textbf{MiniImagenet}~\cite{vinyals2016matching} is sampled from the original ImageNet. There are 100 classes in total, with 600 images for each class in Mini-ImageNet. All images are down-sampled to the size of $84\times 84$. The whole dataset consists of 64 classes for training, 16 classes for validation, and 20 classes for test.
Tasks are extracted from corresponding datasets. It is worthy noted that there are $n$ classes (generally, $n\in \left\{ 5,20 \right\} $) and $k$ training samples (generally, $k\in \left\{ 1,5 \right\} $) per class in a classification task, which is considered as $n$-way, $k$-shot task. The training split samples in a task form a new support set and the remaining validation split samples also form a new query set.

In addition, for supervised learning, we considered several image classification datasets for evaluating with sampled sizes varying from 12500 to 180,000 images. These datasets cover a variety of image classification tasks, including superior-level recognition (ImageNet-1K~\cite{deng2009imagenet}, CIFAR-10) and fine-grained recognition (SVHN). We use the ImageNet dataset for training the SuperNet to get the meta-architecture and use the other two datasets to perform the architecture transfer in supervised learning.
%
%\Subsection{Experiment Settings}
%
%
%The Key Of The \Textbf{S2} Search Space Is Building Blocks That Are Designed Based On A State-Of-The-Art  Manually-Designed  Network Shufflenet V2. The Supernet Backone Of \Textbf{S2} Is Defined As: The Depth Of Super Is Set To 6 Stages; The Number Of Channels In Each Stage Is $C=\Left\{ 16,64,160,320,640,1024 \Right\} $; The Number Of Blocks For Each Stage Is $B=\Left\{ 1, 4, 4, 8, 4, 1 \Right\} $; Stage $\Left\{ 2,3,4,5 \Right\}$ Are Choice Blocks, Where Each Choice Block Has 4 Candidate Building Blocks, Namely Shufflenet Block $3\Times3$, Shufflenet Block $5\Times5$, Shufflenet Block $7\Times7$ And Shufflenet Xception. Only One Micro-Architecture For Each Block In The Searched Network Is Invoked At The Same Time, Which Is Called Single Path. The Size Of The Search Space Is $4^{20}$.

\subsection{AT-NAS For Few-Shot Learning}
According to 5-way 1-shot and 5-way 5-shot settings of few-shot learning, a training task pool and a validation task pool are obtained from the training data and validation data of Mini-ImageNet, respectively.
First, the small SuperNet is the structure of two cells {normal+reduction} for efficiency and the initial channel is set to 16. The SuperNet is trained for 20 epochs with 5000 independent tasks for each epoch. There are $\mathcal{K}$ processes in distributed parallel computing and each process is allocated to $\frac{5000}{\mathcal{K}}$ tasks. The running time is directly reduced by $\mathcal{K}$ times.
The parameter warmup stage is the first 5 epochs. After that, we initialize the population, which maintains 128 different architectures. The SuperNet and the population are optimized on multiple tasks based on Algorithm ~\ref{alg:at-nas_algorithm}.
We use vanilla SGD to optimize the subnets weights with inner-task learning rate $\lambda _{task}=0.01$. In line with Eq.~\ref{eq6}, meta weights are optimized by Adam with a learning rate $\eta _{meta}=0.001$ (initial value).

It is worth noting that we treat the task as the smallest iteration unit instead of the epoch. After every 4 meta weight optimization, the population evolves once for better architectures. Both crossover and mutation occur on population members in the first quarter of the performance rank. In the DARTS search space, each intermediate node in the cell is connected with two previous nodes, which means the corresponding integer string only has two non-zero values. Following the above rule, crossover and mutation are conducted on the corresponding integer string. Each node has a 0.5 ratio for the crossover operation to crossover its connections. For mutation operation, each node is randomly reassigned with a 0.5 ratio. All search and evaluation experiments are performed on 8 NVIDIA 2080Ti and the whole search process takes about 4 GPU days.

\begin{figure*}[htbp]
\centering
\subfigure[Normal cell] {
    \begin{minipage}[t]{0.48\textwidth}
        \centering
        \includegraphics[width=8cm]{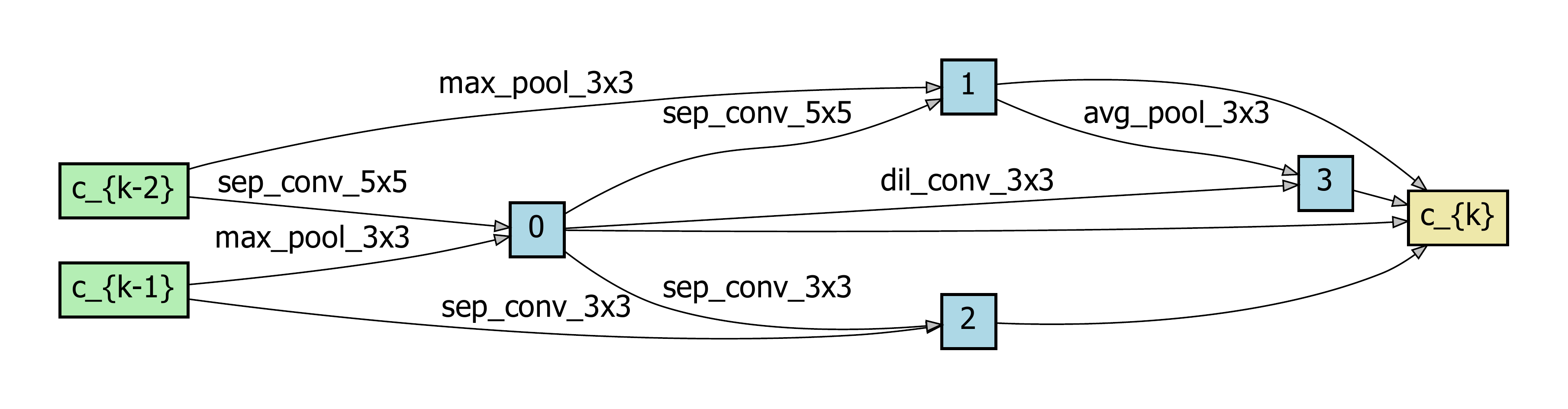}
        \label{fig:normal_cell}
    \end{minipage}
}
\subfigure[Reduction cell] {
    \begin{minipage}[t]{0.48\textwidth}
        \centering
        \includegraphics[width=8cm]{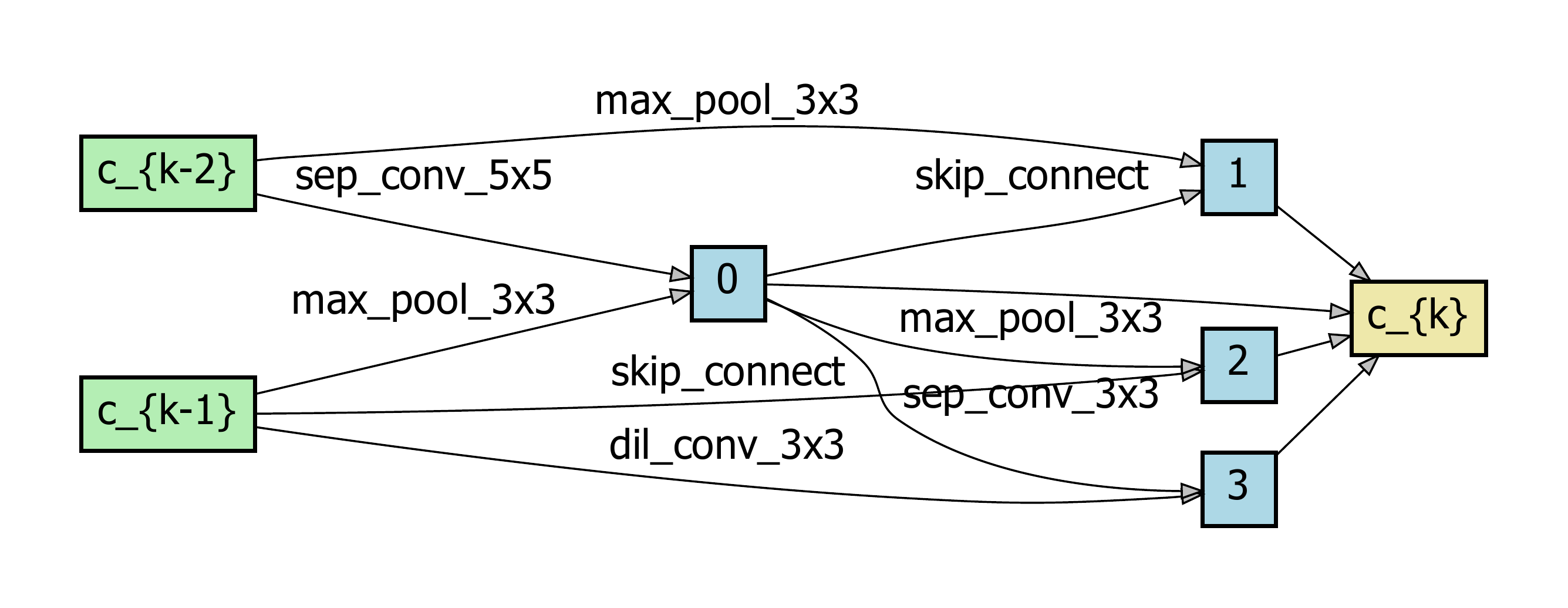}
        \label{fig:reduction_cell}
    \end{minipage}
}
\caption{The best normal and reduction cell found by AT-NAS that are used for the evaluation of 5-way 1-shot tasks}
\label{fig:5-way 1-shot}
\end{figure*}

\begin{figure*}[htbp]
\centering
\subfigure[Normal cell] {
    \begin{minipage}[t]{0.48\textwidth}
        \centering
        \includegraphics[width=8cm]{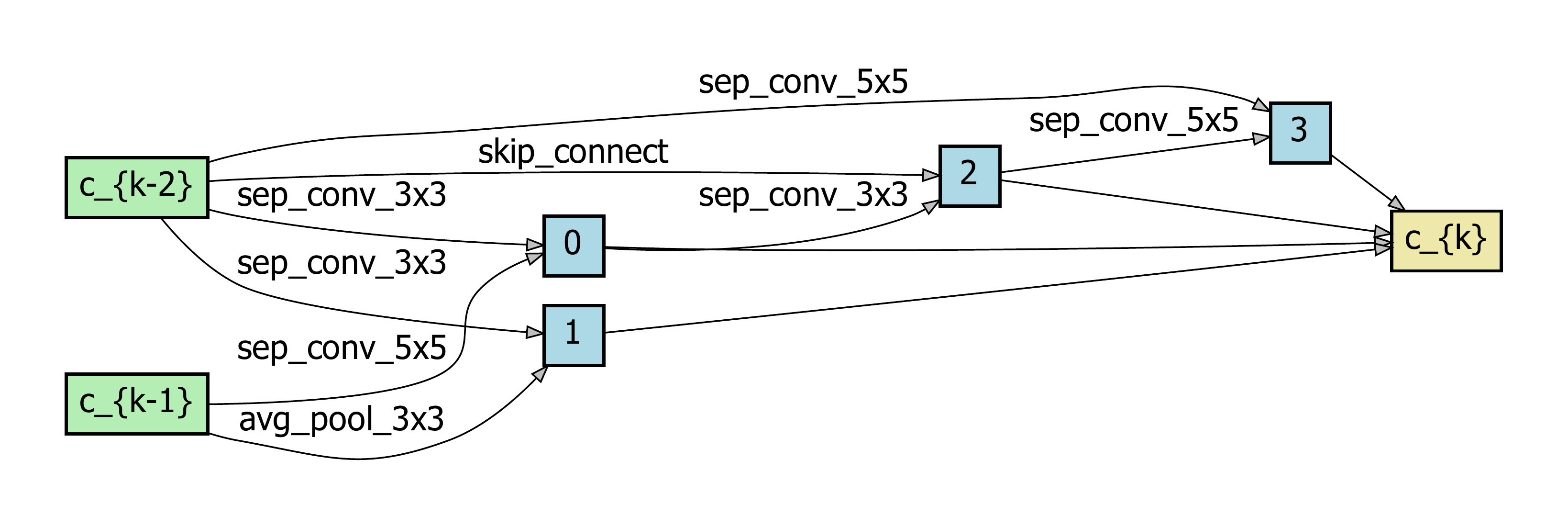}
        \label{fig:normal_cell_55}
    \end{minipage}
}
\subfigure[Reduction cell] {
    \begin{minipage}[t]{0.48\textwidth}
        \centering
        \includegraphics[width=8cm]{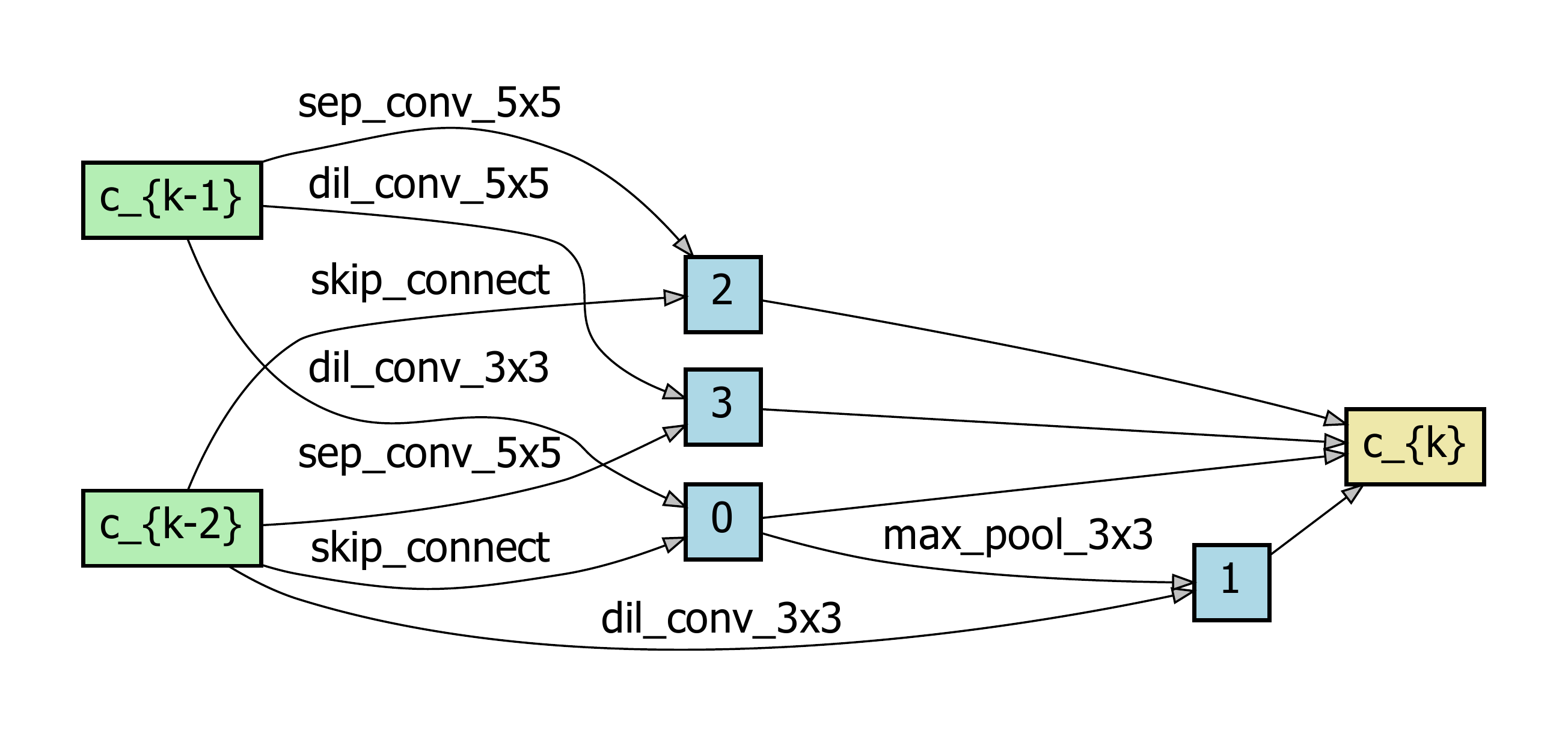}
        \label{fig:reduction_cell_55}
    \end{minipage}
}
\caption{The best normal and reduction cell found by AT-NAS that are used for the evaluation of 5-way 5-shot tasks}
\label{fig:5-way 5-shot}
\end{figure*}

The definitive collection of 128 subnets is seen as the meta-architecture with corresponding weights. After getting the optimal architecture for the specific task $\mathcal{T} _i$ by algorithm~\ref{alg:transfer}, we train the task-specific architecture for 20 epochs with 15000 independent tasks for each epoch. In this process, the weights of subnets are optimized by SGD with $\lambda _{task}=0.1$ and $n=5$ on a task and meta weights are optimized by Adam with $\eta _{meta}=0.001$. The results are summarized in Table~\ref{tab:table2}. AT-NAS surpasses the standard MAML architecture and REPTILE. AT-NAS also outperforms other meta-search methods(e.g., BASE, AutoMAML, T-NAS++) and achieves new state-of-the-art performance. Another outstanding advantage of AT-NAS is to ensure search performance while decoupling weights and architectures, which means that we can explore larger and deeper DARTS model structures. In the few-shot learning experiment, we successfully search for 8-layer architecture with higher performance, and the GPU memory occupied in a single card only increased by 1G. In contrast, T-NAS can not search as out of memory. So AT-NAS can directly find a deeper network architecture without relying on the stack of searched cells.
See Figure~\ref{fig:5-way 1-shot} and Figure~\ref{fig:5-way 5-shot} for the best cells in a 5-way 1-shot task and a 5-way 5-shot task.

\begin{table}
\centering
\caption{5-way accuracy results on Mini-Imagenet. $small$ means 2 cells and $large$ means 8 cells}
\label{tab:table2}
\resizebox{1.0\columnwidth}{!}{
\begin{tabular}{l|cc|cc}
\toprule[2pt]
 \textbf{Methods} & \textbf{Architecture} & \textbf{Params. (K)} & \textbf{5-way, 1-shot} & \textbf{5-way, 5-shot}\\
\hline
 MAML~\cite{finn2017model}             & 4CONV  & 32.9      & 48.70 $\pm $1.84 $\%$  & 63.11 $\pm $ 0.92 $\%$ \\
 FOMAML~\cite{finn2017model}            & 4CONV & 32.9     & 48.07 $\pm $1.75 $\%$  & 63.15 $\pm $ 0.91 $\%$ \\
 MAML++~\cite{antoniou2018train}           & 4CONV  & 32.9      & 52.15 $\pm $0.26 $\%$  & 68.32 $\pm $ 0.44 $\%$ \\
\hline
 Auto-Meta~\cite{kim2018auto}        & Cell   & 28/28        & 49.58 $\pm $0.20 $\%$  & 65.09 $\pm $ 0.24 $\%$ \\
 BASE~\cite{shaw2019meta}             & Cell  & 1200      & -  & 66.20 $\pm $ 0.70 $\%$\\
 M-NAS~\cite{wang2020m}            & Cell   & 27.9      & 51.37 $\pm $1.41 $\%$  & -\\
 T-NAS($small$)~\cite{lian2019towards}     & Cell   & 26.5      & 52.84 $\pm $1.41 $\%$  & 67.88 $\pm $ 0.92 $\%$ \\
 T-NAS++~\cite{lian2019towards}          & Cell   & 26.5      & 54.11 $\pm $1.35 $\%$  & 69.59 $\pm $ 0.85 $\%$   \\
 Auto-MAML~\cite{lian2019towards}        & Cell   & 23.2/26.1      & 51.23 $\pm $1.76 $\%$  & 64.10 $\pm $ 1.12 $\%$ \\
\hline
 \textbf{AT-NAS($small$)}    & \textbf{Cell}   &35.7     & 57.08 $\pm $ 1.24 $\%$   &70.33 $\pm $ 0.64 $\%$ \\
 \textbf{AT-NAS($large$)}    & \textbf{Cell}   &142.8    & 60.92 $\pm $ 0.96 $\%$   &75.56 $\pm $ 0.24$\%$     \\
\bottomrule[2pt]
\end{tabular}}

\end{table}

For Omniglot, the batch size is set to 32 and 16 for 5-way and 20-way classification, respectively. Distributed parallel training allows each process to handle 4 or 2 tasks(assuming there are 8 processes). Distributed training will average the results of 8 processes. The results of AT-NAS on Omniglot are summarized in Table~\ref{tab:table2}. All methods perform similarly well, besides AT-NAS and Meta-SGD having slight advantages over others.

\begin{table}
\centering
\caption{5-way accuracy results on Omniglot dataset}
\label{tab:table3}
\resizebox{1.0\columnwidth}{!}{
\begin{tabular}{l|cc|cc}
\toprule[2pt]
 \textbf{Methods}   & \textbf{5-way, 1-shot} & \textbf{5-way, 5-shot} & \textbf{20-way, 1-shot} & \textbf{20-way, 5-shot}\\
\hline
Matching nets~\cite{vinyals2016matching}             &98.1 $\%$               &98.9 $\%$                &93.8 $\%$               &98.5 $\%$\\
Memory Mod~\cite{kaiser2017learning}                &98.4 $\%$               &99.6 $\%$                &95.0 $\%$               &98.6 $\%$\\
\hline
MAML~\cite{finn2017model}                    & 98.7 $\pm $0.4 $\%$  & 99.9 $\pm $ 0.1 $\%$  & 95.8 $\pm $0.3 $\%$  & 98.9 $\pm $ 0.2 $\%$\\
MAML++~\cite{antoniou2018train}              & 99.47 $\%$  & 99.93 $\%$ & 97.65 $\pm $0.05 $\%$  & 99.33 $\pm $ 0.03 $\%$\\
Meta-SGD~\cite{li2017meta}                                     &99.53 $\pm $0.26 $\%$   & 99.93 $\pm $ 0.09 $\%$  &95.93 $\pm $ 0.38 $\%$  &98.97 $\pm $ 0.19 $\%$\\
REPTILE~\cite{nichol2018first}                                      & 97.68 $\pm $0.04 $\%$  & 99.48 $\pm $ 0.06 $\%$ & 89.43 $\pm $0.14 $\%$  & 97.12 $\pm $
0.32 $\%$ \\
FOMAML~\cite{finn2017model}                  & 98.3 $\pm $0.5 $\%$  & 99.2 $\pm $ 0.2 $\%$ & 89.4 $\pm $0.5 $\%$  & 97.9 $\pm $ 0.1 $\%$ \\
\hline
Auto-Meta~\cite{kim2018auto}                 & 98.94 $\pm $0.07 $\%$  & -                  & -                     & -\\
%BASE~\cite{shaw2019meta}                      & -  & 66.20 $\pm $ 0.70 $\%$\\
M-NAS~\cite{wang2020m}                       & -                     & -                   & 96.2 $\pm $0.16 $\%$  & 99.2 $\pm $0.07 $\%$\\
T-NAS  ~\cite{lian2019towards}               & 99.16 $\pm $0.34 $\%$  & 99.93 $\pm $ 0.07 $\%$   & -                  &- \\
T-NAS++~\cite{lian2019towards}               & 99.35 $\pm $0.32 $\%$  & 99.93 $\pm $ 0.07 $\%$   & -                  &- \\
Auto-MAML~\cite{lian2019towards}             & 98.95 $\pm $0.38 $\%$  & 99.91 $\pm $ 0.09 $\%$   & -                  &- \\
\hline
\textbf{AT-NAS(ours)}                        & 99.38 $\pm $0.30 $\%$  & 99.92 $\pm $0.08 $\%$    & 95.5 $\pm $0.4 $\%$  & 98.9 $\pm $0.07   \\
\bottomrule[2pt]
\end{tabular}}
\end{table}

\subsection{AT-NAS For Supervised Learning}

For presenting the transferability of AT-NAS, a meta-learning-like experiment is conducted to complete across task learning on more large-scale datasets.
Several classes with corresponding images from the ImageNet-1k are randomly sampled by following the settings of the Mini-ImageNet in few-shot learning, where sampled classes are constructed into a 10-way 200-shot 50-query classification task. The number of the sampled classes is the same as the target dataset, which means that we construct many small tasks with similar structures but fewer samples than the target dataset. Unlike few-shot learning, the architecture is allowed to be searched and trained for each task with more sufficient samples. The image size of target datasets is uniformly set to $224 \times 224$ for subsequent transfer. The super network is still a DARTS structure, but we designed three architectures (2-layer, 8-layer) to tap the performance as much as possible.

 A task pool is composed of 500 above-mentioned tasks, of which 450 tasks constitute the training task pool and 50 tasks constitute the validation task pool. Since samples of a single task are enough for the network to extract certain features, we split one task pool into two task pools. One task pool is used for meta weight optimization and the other is used for meta architecture optimization. Namely, meta-weight is updated on the training task pool and meta-architecture is updated on the validation task pool. The initial collection of 1024 architectures is seen as the meta-architecture population to be optimized. The total search process is to run meta-weight optimization and meta-architecture optimization for 50 epochs iteratively. The hyper-parameter settings of the meta-architecture search are consistent with those in AT-NAS for few-shot learning. But the batch size of the sampled architecture is reduced to 16. The sampled subnets are updated for 10 steps on each task. The performance predictor is embedded in the architecture evaluation step of the meta-architecture optimization to save search time.

\begin{figure}[htbp]
\centering
\includegraphics[height=4.6cm]{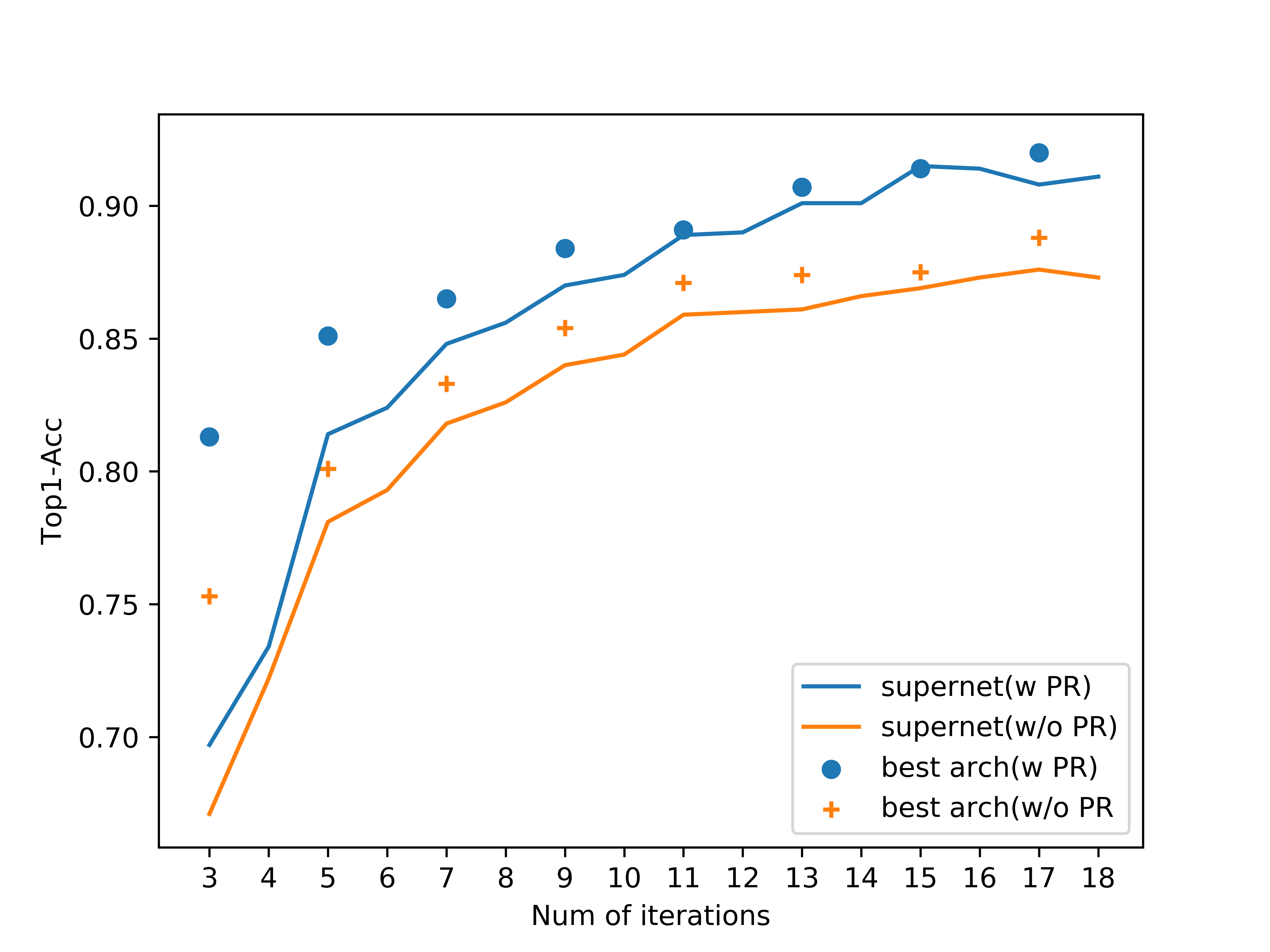}
\caption{The architecture transfer process by evolutionary algorithm with population-halving on CIFAR-10 (Blue)}
\label{fig:train_pr}
\end{figure}

After obtaining the meta-network population that consists of the meta-architecture population and the meta supernet, the entire meta-network population is transferred to the object datasets, such as CIFAR-10, SVHN.  On the target datasets, the meta supernet is trained for 20 epochs. The population is updated every 2 epochs for a total of 10 updates. The meta-network population is seen as the initial population, and each evolution is carried out in light of the population-halving strategy.
%The architecture transferred to multiple datasets by AT-NAS can be found in Figure~\ref{fig:transfer} and Algorithm~\ref{alg:transfer}.
The architecture transferred to multiple datasets by AT-NAS can be found in Algorithm~\ref{alg:transfer}.
The transfer process is shown in Figure~\ref{fig:train_pr}.

\textbf{Performance on CIFAR10 Dataset.} The result of AT-NAS on CIFAR10 can be found in Table ~\ref{tab:table4}. AT-NAS (Multi-Task) is sampled from meta-population without further fine-tuning. The model is used as a baseline since it does not access any information on the object dataset. AT-NAS (CIFAR-10 Tuned) is the model derived from the meta-network fine-tuned on CIFAR-10. Not surprisingly, the model performs best as it has access to both similar information of multiple tasks and unique information of the target dataset. Compared with the accuracy index, AT-NAS has more obvious advantages in search time. For AT-NAS (CIFAR-10 Tuned), we only finetune the meta-network for 0.05 GPU days. The required time is significantly less than that required for searching from scratch of the baseline NAS algorithms. Concerning FLOPs, the proposed models are comparable to the baseline models but have higher performance.

\begin{table}
\centering
\caption{Classification Accuracies on CIFAR10}
\label{tab:table4}
\resizebox{1.0\columnwidth}{!}{
\begin{tabular}{l|cc|c|c}
\toprule[2pt]
 \textbf{Model} & \makecell[l]{\textbf{Search} \\ \textbf{Method}} & \makecell[l]{\textbf{Params.} \\ \textbf{(M)} } & \makecell[l]{\textbf{Top1 Test} \\ \textbf{Error($\%$)}} & \makecell[l]{\textbf{Search Time} \\ \textbf{(GPU-days)}}\\
\hline
 NASNet-A+cutout~\cite{zoph2016neural}      & RL               & 3.3      &2.65 $\%$  & 1800\\
 AmoebaNet-A+cutout~\cite{real2019regularized}   & EA               & 3.2      &3.34 $\pm$ 0.06 $\%$  & 3150\\
 PNAS~\cite{liu2018progressive}                 & SBMO             & 3.2      &3.41 $\pm$ 0.09 $\%$  & 225\\
 ENAS+cutout~\cite{pham2018efficient}                 & RL        & 4.6      &2.89 $\%$  & 0.5\\
\hline
 DARTS(1st order)+cutout~\cite{liu2018darts}     & gradient  & 3.3      &3.00 $\pm$ 0.14 $\%$  & 4\\
 DARTS(2nd order)+cutout~\cite{liu2018darts}     & gradient  & 3.3      &2.76 $\pm$ 0.09 $\%$  & 4\\
 PC-DARTS+cutout~\cite{xu2019pc}             & gradient  & 3.6      &2.57 $\pm$ 0.07 $\%$  & 3.8\\
 T-NAS(first-order) + cutout ~\cite{lian2019towards}              & gradient         & 3.4      &2.98 $\pm$ 0.12 $\%$      & 0.043 Adap\\
 BASE(Multi-task Prior)~\cite{elsken2020meta} & gradient         & 3.2      &3.18 $\%$  & 0.04 Adap / 8 Meta\\
 BASE(CIFAR-10 Tuned)~\cite{elsken2020meta} & gradient           & 3.1      &2.83 $\%$  & 0.04 Adap / 8 Meta\\
\hline
 CARS~\cite{yang2020cars}                 & EA               & 3.6          & 2.62$\%$      & 0.14\\
 \textbf{AT-NAS-A(Multi-task Prior)}      & EA               & 3.2          & 3.10 $\pm$ 0.16 $\%$     & -\\
 \textbf{AT-NAS-A(CIFAR10 Tuned)}        & EA               & 3.1          & 2.81 $\pm$ 0.12 $\%$     & 0.05 Adap \\
 \textbf{AT-NAS-B(Multi-task Prior)}      & EA               & 11.0         & 2.89 $\pm$ 0.13 $\%$     & -\\
 \textbf{AT-NAS-B(CIFAR10 Tuned)}        & EA               & 10.8         & 2.63 $\pm$ 0.10 $\%$     & 0.06 Adap \\
\bottomrule[2pt]
\end{tabular}}

\end{table}

\textbf{Performance on SVHN Dataset}. The result of our Meta Architecture Search on SVHN is shown in Table ~\ref{tab:table5}. Since SVHN has the same number of classes as CIFAR10, we use the same multi-task prior that trained on the Imagenet datasets and quickly adapt the meta-network to SVHN in less than an hour. We also train the CIFAR10 specialized architecture searched by DARTS. AT-NAS-B(CIFAR10 Tuned)achieves the best performance in our experiments. AT-NAS-A(CIFAR10 Tuned) has comparable performance to other work for a similar model size.

\begin{table}
\centering
\caption{Classification Accuracies on SVHN}
\label{tab:table5}
\resizebox{1.0\columnwidth}{!}{
\begin{tabular}{l|c|c|c}
\toprule[2pt]
 \textbf{Model} & \makecell[l]{\textbf{Params.} \\ \textbf{(M)} } & \makecell[l]{\textbf{Top1 Test} \\ \textbf{Error($\%$)}} & \makecell[l]{\textbf{Search Time} \\ \textbf{(GPU-days)}}\\
\hline
 WideResnet~\cite{zagoruyko2016wide}                                      &11.7               & 1.30 $\pm $ 0.03$\%$     & -\\
 MetaQNN~\cite{baker2016designing}                                         &9.8                & 2.24 $\%$                 & 100\\
\hline
 DARTS(CIFAR10 Searched)~\cite{liu2018darts}     &3.3                &2.09  $\%$             & 4\\
 BASE(Multi-task Prior)~\cite{elsken2020meta}    &3.2                &2.13  $\%$             & 0.04 Adap \\
 BASE(SVHN Tuned)~\cite{elsken2020meta}          &3.2                &2.01  $\%$            & 0.04 Adap \\
\hline
 \textbf{AT-NAS-A(Multi-task Prior)}             &3.2                &2.14 $\pm $ 0.10 $\%$     &-\\
 \textbf{AT-NAS-A(SVHN Tuned)}                   &3.1                &2.02 $\pm $ 0.08 $\%$     & 0.05 Adap \\
 \textbf{AT-NAS-B(Multi-task Prior)}            &10.7                &2.05 $\pm $ 0.08 $\%$     &-\\%&3.10 $\pm$ 0.16      & 8 Meta\\
 \textbf{AT-NAS-B(SVHN Tuned)}                   &10.5               &1.88 $\pm $ 0.06 $\%$     & 0.06 Adap \\%&2.80 $\pm$ 0.11      &0.05 Adap / 8 Meta\\
\bottomrule[2pt]
\end{tabular}}

\end{table}

\section{Ablation Study}

\subsection{Accuracy Predictor Performance}
\begin{figure}[htbp]
\centering
\includegraphics[height=4.6cm]{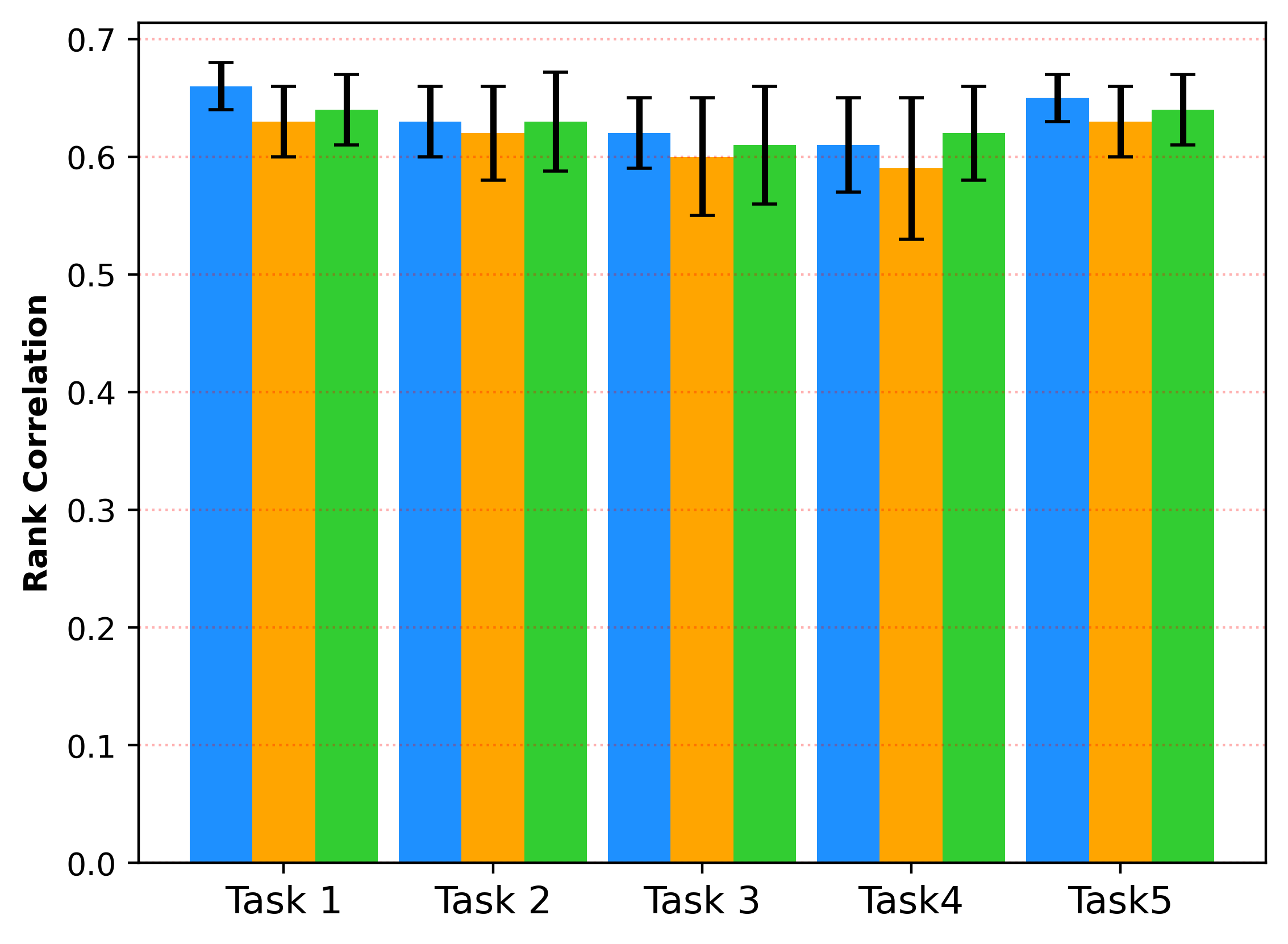}
\caption{Rank correlation between predicted accuracy and true accuracy of different proxy models across different tasks}
\label{fig:correlation tasks}
\end{figure}

In this subsection, we evaluate the effectiveness of different performance predictor models. 350 architectures are uniformly sampled from the search space and trained using MAML(first-order) for 100 epochs on 20 tasks with a 10-way 200-shot 50-query setting. Each of them finetunes for 30 epochs on the other 5 tasks different from the above tasks. All tasks are sampled from ImageNet. From the 300 pairs of architectures and accuracy obtained on each task, we remain 50 pairs available for training the predictor models and 20 pairs as a subset for testing. Figure~\ref{fig:correlation tasks} shows the Spearman rank correlation between predicted and factual accuracy of different proxy predictors among 5 tasks. The number of categories is the same, but the specific categories are different. From the trade-off view of minimizing the number of training samples and maximizing the prediction accuracy, Random Forest is chosen as an accuracy predictor embedded in evaluating architectures of AT-NAS in supervised learning experiments.

In addition, we compared the performance difference between AT-NAS with predictor and AT-NAS without predictor. From Table~\ref{tab:table7}, the AT-NAS with predictor has almost no performance loss, and the search time is significantly reduced.

%\begin{figure}[htbp]
%\begin{center}
%\includegraphics[width=8cm]{correlation_tasks.png}
%\end{center}
%\caption(Rank correlation between predicted accuracy and true accuracy of different proxy models across different tasks. }
%\label{fig:correlation tasks}
%\end{figure}

\begin{table}
\centering
\caption{Comparison of accuracy before and after embedding the performance predictor on CIFAR-10}
\label{tab:table7}
\begin{tabular}{l|c|c}
\toprule[1pt]
Methods                            &\makecell[l]{Top1 Test \\ Error($\%$)}          & \makecell[l]{Search Time \\ (GPU Days) } \\ %& Flops(G) \\
\hline
ATNAS                              &2.45 $\pm$ 0.10 $\%$      &10   \\ %& 15.64   \\
\hline
ATNAS(predictor)                   &2.81 $\pm$ 0.12 $\%$      &4   \\ %& 9.74      \\
\bottomrule[1pt]
\end{tabular}
\end{table}

\begin{table}
\centering
\caption{The accuracy and Search Time of 1,4,8 iterations. Meta task batch size is set to 4}
\label{tab:table8}
\begin{tabular}{l|c|c}
\toprule[1pt]
Iterations                            &Accuracy          & \makecell[l]{Search Time \\ (GPU Days) } \\ %& Flops(G) \\
\hline
1                                     &54.60 $\pm $ 1.29 $\%$    &12   \\ %& 15.64   \\
\hline
4                                     &54.23 $\pm $ 1.36 $\%$    &4.2   \\ %& 9.74      \\
\hline
8                                     &53.25 $\pm $ 1.40 $\%$    &3.5   \\
\bottomrule[1pt]
\end{tabular}

\end{table}

\begin{table}
\centering
\caption{The accuracy and Search Time of 1,2,4 meta task batch size. The iteration number is set to 4}
\label{tab:table9}
\begin{tabular}{l|c|c}
\toprule[1pt]
Meta Task Batch Size                            &Accuracy          & \makecell[l]{Search Time \\ (GPU Days) } \\ %& Flops(G) \\
\hline
1 task                                          &54.60 $\pm $ 1.29 $\%$                &14   \\ %& 15.64   \\
\hline
2 tasks                                          &57.08 $\pm $ 1.13 $\%$               &4.2   \\ %& 9.74      \\
\hline
4 tasks                                          &53.65 $\pm $ 1.37 $\%$               &2.9   \\
\bottomrule[1pt]
\end{tabular}
\end{table}

\subsection{Analysis of search Hyperparameters}
The search process of AT-NAS is divided into meta parameters optimization and meta architecture optimization. The two processes are separate but run iteratively to affect the final searched network. The latter runs once after the former $k$ times, thus the iteration number $k$ is a very important hyperparameter. Table~\ref{tab:table8} reflects that iterations are negatively correlated with search accuracy and positively correlated with search time.
In addition,  the respective meta task batch size within each process will also affect the final network performance in distributed parallel meta-learning, as AT-NAS learns the information of $B_{meta}\times \mathcal{K}$ tasks at a time.
Table~\ref{tab:table9} shows that meta batch size and iterations have similar properties.
At the same time, the meta task batch size has a certain correlation with the meta learning rate.
The parameter sweep experiments are conducted in the few-shot learning experiment for identifying the optimal search hyperparameter values.

%补超参数表

\section{Conclusion}
This paper investigates an efficient algorithm to search a meta-network over multiple tasks. To reduce the time for the network to adapt a new task, we present an across-task NAS framework that fully intergrates the concept of MAML and EA-based NAS with weight sharing, namely AT-NAS. The searched meta-network is able to adapt to a new task quickly based on a task-agnostic representation extracted from existing tasks, which is more flexible and faster than ordinary NAS algorithms. We also show a distributed parallel MAML algorithm and how to use an online regressor as a proxy model to predict the accuracy of subnets. Experimental results on standard few-shot learning benchmarks show that the model derived from AT-NAS is on-par or better than other similar meta architecture search methods. As for the supervised learning, the transfer architecture generated by AT-NAS achieves comparable performance to other baselines but with  $ 60 \times $ less search cost on the target dataset. This demonstrates the possible great efficiency gains by the meta architecture search over existed task distributions. Since the framework of AT-NAS has no concern with specific meta-learning approaches as well as architecture search methods, we plan to extend the AT-NAS beyond classification to other multiple task.

\begin{acknowledgements}
We gratefully acknowledge the support of National Key R\&D Program of China (2018YFB1308400) and Natural Science Foundation of Zhejiang Province (NO. LY21F030018).
\end{acknowledgements}


%\bibliographystyle{plain}
\bibliographystyle{spmpsci}      % mathematics and physical sciences
\bibliography{ijcai21_norm}   % name your BibTeX data base

%\begin{thebibliography}{}
%%
%% and use \bibitem to create references. Consult the Instructions
%% for authors for reference list style.
%\end{thebibliography}

\end{document}